\documentclass{article}
\PassOptionsToPackage{numbers}{natbib}
\usepackage[preprint]{neurips_2020}

\usepackage{amssymb, amsmath, amsthm, comment, xspace, enumerate}

\usepackage[utf8]{inputenc} 
\usepackage[T1]{fontenc}    
\usepackage{hyperref}       
\usepackage{url}            
\usepackage{booktabs}       
\usepackage{amsfonts}       
\usepackage{nicefrac}       
\usepackage{microtype}      
\usepackage{graphicx,subfigure}

\usepackage[mathcal]{eucal}

\DeclareMathOperator*{\expectation}{\mathbb{E}}

\DeclareMathOperator*{\argmax}{\rm{argmax}}

\def\<{\begin{equation}}
\def\>{\end{equation}}

\def\b{\alpha}

\newtheorem{thm}{Theorem}
\newtheorem{remark}[thm]{Remark}

\newtheorem{thm2}{Theorem}
\newtheorem{lemma}[thm2]{Lemma}

\newtheorem{thm3}{Theorem}
\newtheorem{theorem}[thm3]{Theorem}

\newtheorem{thm4}{Theorem}
\newtheorem{definition}[thm4]{Definition}

\title{Learning and Inference in Imaginary Noise Models}

\author{%
  Saeed Saremi \\
Redwood Center for Theoretical Neuroscience\\
       UC Berkeley, CA 94720-3198, USA\\
       NNAISENSE Inc., Austin, TX\\
  \texttt{saeed@berkeley.edu} \\
}

\begin{document}

\maketitle

\begin{abstract}
Inspired by recent developments in learning smoothed densities with empirical Bayes, we study variational autoencoders with a decoder that is tailored for the random variable $Y=X+N(0,\sigma^2 I_d)$. A notion of smoothed variational inference emerges, where the smoothing is implicitly enforced by the noise model of the decoder; ``implicit'', since during training the encoder only sees clean samples. This is the concept of imaginary noise model, where the noise model dictates the functional form of the variational lower bound $\mathcal{L}(\sigma)$, but the noisy data are never seen during learning. The model is named $\sigma$-VAE. We prove that all $\sigma$-VAEs are equivalent to each other via a simple $\beta$-VAE expansion: $\mathcal{L}(\sigma_2) \equiv \mathcal{L}(\sigma_1,\beta)$, where $\beta=\sigma_2^2/\sigma_1^2$. We prove a similar result for the Laplace distribution in exponential families. Empirically, we report an intriguing power law $\mathcal{D}_{\rm KL} \sim \sigma^{-\nu}$ for the learned models and we study the inference in the $\sigma$-VAE for unseen noisy data. The experiments were performed on MNIST, where we show that quite remarkably the model can  make reasonable inferences on extremely noisy samples even though it has not seen any during training. The vanilla VAE completely breaks down in this regime. We finish with a hypothesis (the XYZ hypothesis) on the findings here.\end{abstract}

\section{Introduction}
This work was motivated by developing a notion of \emph{smoothed} variational inference in the framework of variational autoencoders~\citep{kingma2013auto,rezende2014stochastic} which was particularly inspired by the recent progress in learning smoothed densities with empirical Bayes~\cite{saremi2019neural}. The notion of smoothed variational inference can also be motivated from the perspective of robust inference as there is a clear connection between smoothness and robustness~\cite{cohen2019certified}. From the angle of robustness, we would like the variational inference to be robust to noise, where we may consider the isotropic Gaussian $N(0,\sigma^2 I_d)$ as the noise model and we may be interested from the outset in robustness to large amounts of noise. From the angle of smoothing, we would like to formulate the problem of variational inference for the random variable $Y = X+N(0,\sigma^2 I_d),$
even though we usually\textemdash but not always~\cite{robbins1956empirical}\textemdash start  with the i.i.d. sequence $x_1,\dots, x_n$. 

\begin{figure}[t!]
 \begin{center}
 \begin{subfigure}[$X$]{\includegraphics[width=0.3\textwidth]{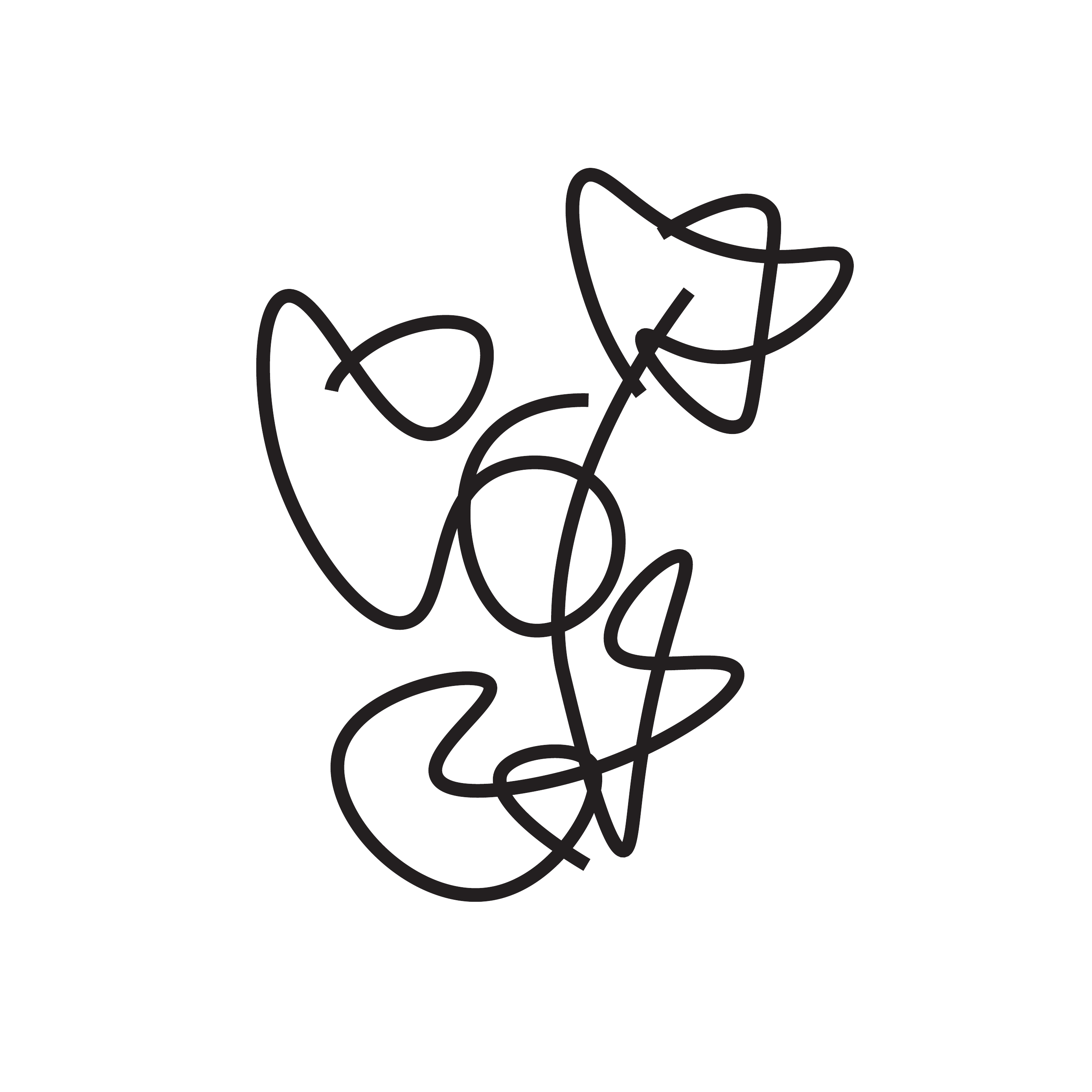}}
 \end{subfigure}
 \begin{subfigure}[$Y$]{\includegraphics[width=0.3\textwidth]{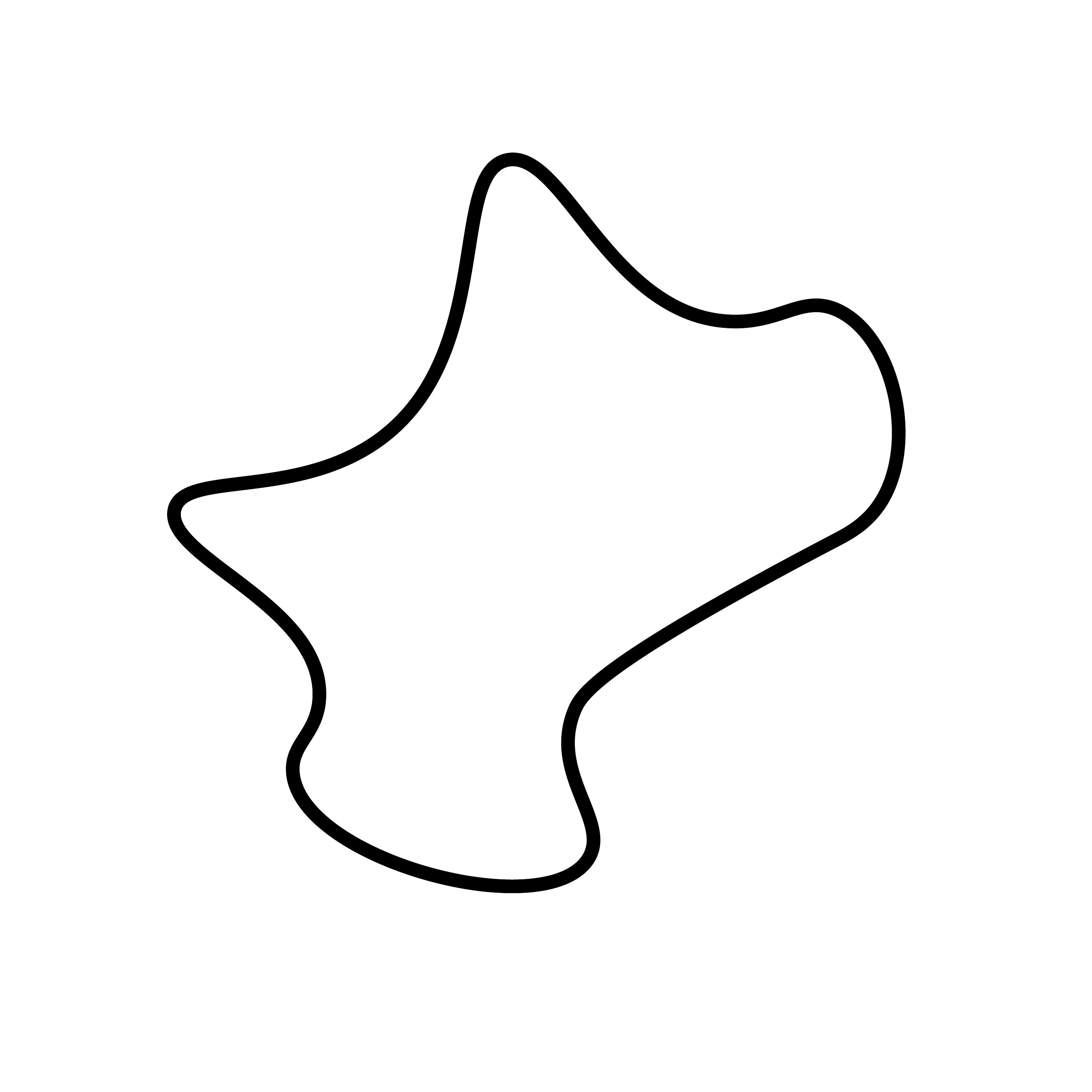}}
  \end{subfigure}	
  \begin{subfigure}[$Z$]{\includegraphics[width=0.3\textwidth]{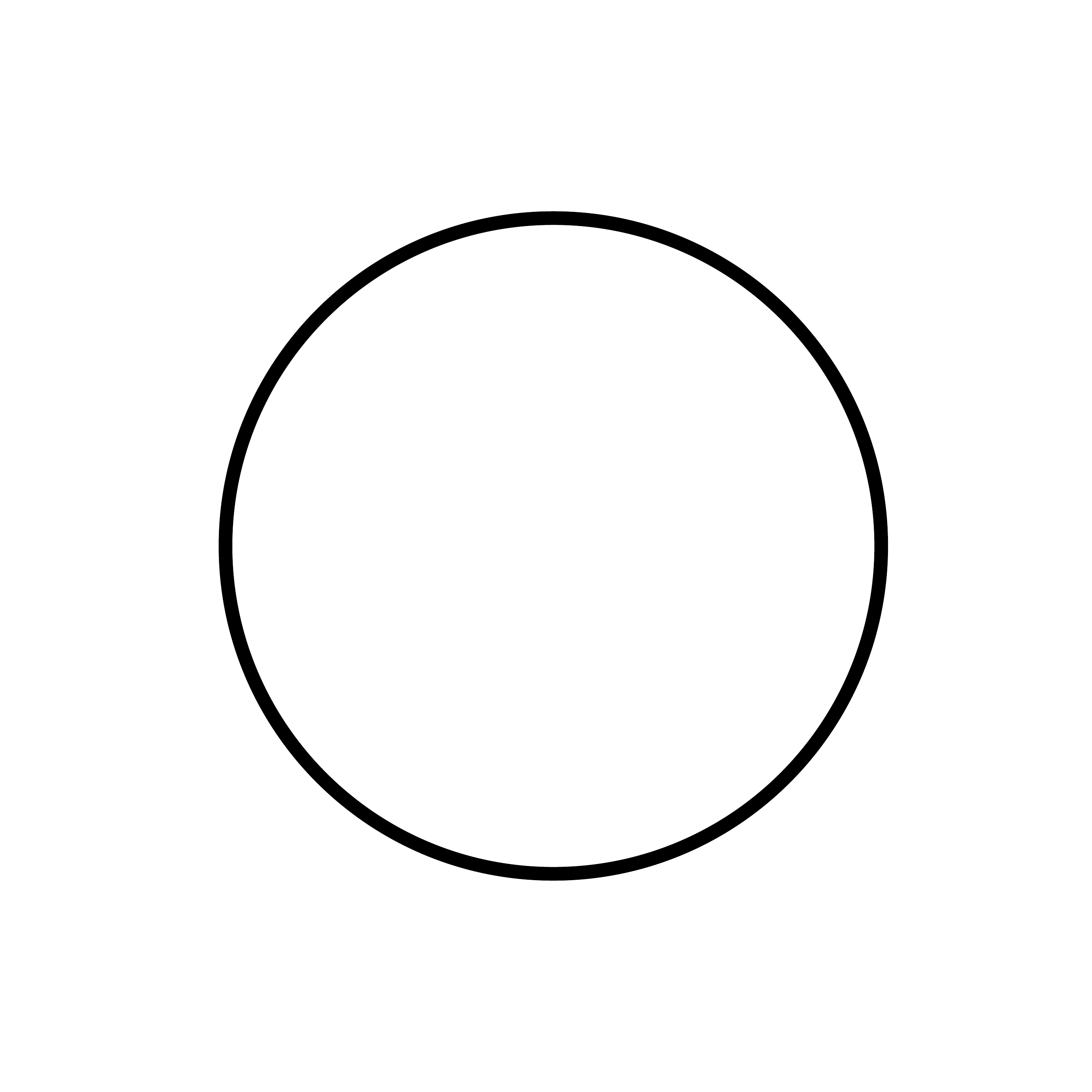}}
 \end{subfigure}
 \caption{({\it The XYZ hypothesis}) (a) $X$ is complex  when viewed in the ambient space $\mathbb{R}^d$. We yet to have a clear formulation of the notion of manifold for data distributions, but in this schematic the manifold is visualized as ``mixture of mess''.  (b) $Y=X+N(0,\sigma^2 I_d)$ is pictured here as a true smooth manifold. In essence, we view $Y$ as $X$ \emph{disintegrating-expanding}~\citep{saremi2019neural} by adding samples from the Gaussian $\approx \text{Unif}(\sigma \sqrt{d} S_{d-1})$ to $X$. (c) $Z$ is pictured assuming $d_z$ itself being relatively high. The $\sigma$-VAE's noise model is defined by $Y$, but that is only imaginary in that the model only sees clean samples from $X$. The XYZ hypothesis states that the approximate posterior inference over $Z$ in $\sigma$-VAE becomes smoothed and more robust as a result of this imaginary noise model of the world.}
 \label{fig:XYZ}
 \end{center}
 \end{figure}

This wish for formulating learning and inference in terms of the random variable $Y = X+N(0,\sigma^2 I_d)$ is somewhat grounded and goes beyond our strong desire for (more) smoothness/robustness:\begin{itemize}
	\item The concentration of $X$ in the ambient space $\mathbb{R}^d$ is almost always very complex and one typically starts with an assumption on the existence of a data manifold~\cite{bengio2013representation}. Indeed, this assumption has had deep impacts in framing the problem of dimensionality reduction in machine learning~\cite{roweis2000nonlinear, tenenbaum2000global, saul2003think}. However, one can also argue that this is not a good starting point in high dimensions.  In latent variable models, the manifold assumption shows itself in disguise in \emph{assuming} latent spaces with small number of dimensions $d_z \ll d$. In this paper, we aim to move away from this assumption in formulating smoothed variational inference.
	\item For $Y$, the notion of a smooth manifold could be realized in high dimensions due to the concentration of $N(0,\sigma^2 I_d) \approx \text{Unif}(\sigma \sqrt{d} S_{d-1})$ where geometrically it has the effect of mapping data points to high-dimensional spheres (see Fig.~\ref{fig:XYZ}). The concentration of measure phenomenon~\cite{ledoux2001concentration, tao2012topics} and its impacts in high dimensions was analyzed for a toy example in~\cite{saremi2019neural} where they characterized the ``disintegration-expansion'' effect analytically. An intuition develops with the takeaway that the concentration of $Y$ is quite different than $X$, much smoother, with an effective dimension which is of the order of $d$ of the ambient space.
\end{itemize}

This paper is organized as follows. In Sec.~\ref{sec:VAE} we introduce variational autoencoders~\citep{kingma2013auto,rezende2014stochastic}. In Sec.~\ref{sec:deen}  we discuss neural empirical Bayes~\citep{saremi2019neural} which shares goals with VAEs, with  strengths for some problems and with fundamental limitations for others. The shortcomings become a motivation for bringing smoothed density/energy models and variational autoencoders closer together. In Sec.~\ref{sec:SVAE} we present our main contributions in formalizing the notion of \emph{smoothed variational inference} starting with the definition of \emph{imaginary noise models} centered around the decoder of variational autoencoders:\begin{quote}The decoder is originally written with $Y=X+N(0,\sigma^2 I_d)$ in mind, but we use a decoder for $X$ with the same  functional form, imagining that $X$ is (very) noisy. The decoder is unrealistic! We essentially make use of the freedom we have in latent variable models to write \emph{any} model for the joint density and here we tailor it towards $Y$, arriving at the ELBO $\mathcal{L}(\sigma)$ which has a simple but very important  dependence on $\sigma$. It is the job of the inference network to make sense of this imaginary noise model of the world and we show empirically to have the effect of smoothing the variational inference, and making it more robust to noise. It also has a related effect of bringing the posterior closer to the  prior with the power law $\mathcal{D}_{\rm KL} \sim \sigma^{-\nu}$ ($\nu\approx 1.15$) which summarizes in an algebraic form the \emph{hypothesis} that the inference is smoother for larger $\sigma$ (see Fig.~\ref{fig:XYZ}). The model is named $\sigma$-VAE.\end{quote} In Sec.~\ref{sec:experiments} we present experiments on MNIST. We especially showcase the results for $\sigma=0.9$, pushing the model to its very limits. The MNIST database~\cite{lecun1998gradient} is  now considered too simple, but in our view the problem of robust inference on the handwritten digits with very large amounts of noise is a good new challenge, especially since we lack a formal notion of robust/smoothed variational inference. We demonstrate that the inference in $\sigma$-VAE is remarkably robust to noise even though the model does not see noisy data during training. This even holds for high levels of salt-and-pepper noise.  In Sec.~\ref{sec:related} we trace the motivations of this work in the literature on robust classification and also discuss other findings in the literature on variational autoencoders. We finish the paper elaborating more on the XYZ hypothesis described briefly in Figure~\ref{fig:XYZ} and with some discussions on the findings of this study.

\begin{remark} Upon completion of this work, we discovered some rather deep connections to $\beta$-VAE~\cite{higgins2017beta} encapsulated in Theorem~\ref{thm:sigmavae} which draws a formal link between smoothing the variational inference and learning disentangled representations. In its summary, the $\beta$ ``pops up'' in mapping between different $\sigma$-VAE models. Therefore, from the unification perspective, one could view $\sigma$-VAEs as  more fundamental since any $\beta$-VAE in this model can be easily mapped to a $\sigma$-VAE with $\beta=1$. \end{remark}

\section{Auto-Encoding Variational Bayes} \label{sec:VAE} Consider the random variable $X$ in $\mathbb{R}^d$ in the context of latent variable models, where we introduce the (latent) random variable $Z$ in $\mathbb{R}^{d_z}$ with a parametrized joint density $p_\theta(x,z)$ and our goal is to learn $\theta$ such that $p_\theta(x) = \int p_\theta(x,z) dz$ is a good \emph{approximation} to $p(x)$. Taking Kullback-Leibler divergence as the metric of choice to measure the approximation, and given the i.i.d. sequence $x_1,\dots,x_n$, the problem of learning $\theta$ is then formulated by maximizing the log-likelihood: $ \mathcal{L}(\theta) = \sum_i \log p_\theta(x_i)$. In directed graphical models, one takes another leap of faith and assumes that the directed factorization
$$ p_\theta(x,z) = p_\theta(x|z) p_\theta(z),$$
is a good model for  $X$. How ``good'' this model is clearly depends on $X$ and how it was generated. For example, if underlying $X$ is an Ising model with higher order interactions, then one is better off doing learning and inference in a \emph{Boltzmann machine}~\cite{hinton1986learning}. (Un)fortunately, we almost never have \emph{a priori} knowledge of $X$ and there is tremendous value in developing general purpose inference and learning frameworks for latent variable models. The framework of \emph{variational inference} for directed graphical models~\cite{jordan1999introduction}, with roots in \emph{mean field methods} in statistical mechanics~\cite{anderson1987mean}, has grown as a strong candidate for such a general-purpose machinery. To motivate the approach, start with
$$ \log p_\theta(x) =  \log \int p_\theta(x|z) p_\theta(z) dz.$$
There are advanced MCMC methods~\cite{mackay2003information} to find good estimates of the integral, but for \emph{learning} $\theta$ the integral must be estimated at each step of the optimization procedure which is intractable to be used in a general purpose framework. In variational inference, one approaches this problem by studying \emph{another} intractable problem\textemdash approximating the posterior $p_\theta(z|x)$:
$$ q_\phi(z|x) \approx p_\theta(z|x).$$
Indeed, approximating the posterior is also intractable due to Bayes: $p_\theta(z|x)=p_\theta(x,z)/p_\theta(x)$. In other words, in probabilistic graphical models, the problem of modeling $X$ and the problem of the posterior inference over $Z$ are \emph{duals} and the complexity of the two problems ``match'' in some loose sense. (This duality comes up again briefly in the discussion of the XYZ hypothesis at the end.) This duality in mind, in variational inference one opts for approximating $p_\theta(z|x)$. Taking a flexible yet tractable $q_\phi(z|x)$ as the candidate, we derive a lower bound for $\log p_\theta(x)$ using Jensen's inequality:
\< \label{eq:elbo0} \log p_\theta(x) \geq \mathbb{E}_{q_\phi(z|x)} \log p_\theta(x|z) - \mathcal{D}_{\rm KL}[q_\phi(z|x),p_\theta(z)], \>
where $\mathcal{D}_{\rm KL}$ is the Kullback-Leibler divergence measuring how far the prior is from the posterior, and the first term measures the reconstruction performance of the autoencoder i.e. how good the generative network is as measured by the inference network. The right hand side in the inequality is referred to by \emph{evidence lower bound} (ELBO) denoted by $\mathcal{L}(x,\theta,\phi).$ This is our starting point where the framework of choice for learning and inference is the variational autoencoder (VAE)~\cite{kingma2013auto,rezende2014stochastic} with the important invention of the reparameterization trick that was developed to pass gradients through noise crucial to having low variance estimates for $\nabla_\phi \mathcal{L}$ and  $\nabla_\theta \mathcal{L}$. Having low variance estimates for the gradients is a must in scaling the variational inference to high dimensions and large datasets.

\section{Neural Empirical Bayes} \label{sec:deen}
Empirical Bayes as formulated by Herbert Robbins~\cite{robbins1956empirical} is one of the most influential works in statistics~\cite{robbins2012selected}. There, the starting point is \emph{not} the i.i.d. samples from $X$ but from $Y$. The first observation is that given a noisy measurement $Y=y$, the least-squares estimator of $X$ is the Bayes estimator. What is quite surprising though is that the Bayes estimator can be written \emph{purely} based on the density of $Y$. The only requirement is to know the measurement kernel $p(y|x)$. For the isotropic Gaussian kernel, the Bayes estimator $\widehat{x}(y) = \expectation[X|y]$ is given by~\cite{miyasawa1961empirical} (see~\cite{raphan2011least} for a review): 
$$ \widehat{x}(y) = y + \sigma^2 \nabla \log p(y).$$
Neural empirical Bayes~\cite{saremi2019neural} is based on corrupting the i.i.d. samples $x_1,\dots,x_n$ from $X$ and generating samples $y_{ij}=x_i+\varepsilon_j$ from $Y$ which is then given to the ``experimenter'' in the school of Robbins~\cite{robbins1956empirical} which aims at learning the density of $Y$. The learning algorithm is set up by parametrizing  the \emph{energy function}\textemdash not the score function~\cite{saremi2019approximating}\textemdash of $Y$ with a neural network denoted by $f: \mathbb{R}^d \rightarrow \mathbb{R}$ with parameters $\vartheta$: $-f_\vartheta(y) = \log p_{\vartheta} (y) + \log Z(\vartheta)$, where  $Z(\cdot)$ is the \emph{partition function} which drops out from the learning objective $\mathcal{L}(\vartheta)$:
\< \label{eq:deen} \mathcal{L}(\vartheta) = \mathbb{E}_{x,y} \Vert x-\widehat{x}_\vartheta(y) \Vert^2,\>
\< \label{eq:xhat} \widehat{x}_\vartheta(y) = y-\sigma^2 \nabla f_\vartheta(y). \>
In summary, neural empirical Bayes\textemdash with the birth name DEEN~\cite{saremi2018deep}\textemdash is designed around learning the \emph{unnormalized}~\cite{hyvarinen2005estimation} density of $Y$, and for that problem it is much more efficient than variational autoencoders since the energy is computed \emph{deterministically} by the neural network. But that is ultimately its biggest weakness: the absence of inference and the lack of a latent space $\approx$ a mind~\cite{tenenbaum2011grow}.


\section{Smoothed Variational Inference}\label{sec:SVAE} Setting up a latent variable model for $Y$ appears to be straightforward. The first step is to set up the joint density $p_\theta(y,z)=p_\theta(y|z) p_\theta(z)$. Since $Y=X+N(0,\sigma^2 I_d)$, the generative model is already ``in front of us'' and the conditional density is given by \< \label{eq:noisemodel} p_\theta(y|z) = \mathcal{N}(y|\mu_y(z,\theta),\sigma^2 I_d).\> One can easily derive the ELBO after choosing an approximate posterior $q_\phi(z|y)$, but the problem is that in learning  $(\theta,\phi)$  we only see noisy samples. There is therefore a very big difference with DEEN from the very beginning. There, in the learning objective (Equation~\ref{eq:deen}) the expectation is over the joint $(x,y)$. Here, only $y$ is left. We did wish to have a smoothed variational inference but approximating the posterior $p_\theta(z|y)$ by \emph{only} observing very noisy samples is a recipe for disaster.\footnote{We ran experiments to confirm this! In addition, see~\cite{im2017denoising} which also breaks down in the large  noise regime.} We propose an alternative for formulating the notion of smoothed variational inference: 
\begin{quote}The idea is to \emph{imagine} that $X$ itself is noisy, i.e. our world model is that everything we measure is very noisy as\textemdash having robust inference in mind\textemdash we cannot trust the world.\footnote{``We cannot trust the world'' is in the context of a world \emph{without} adversaries (more on that later). It is a colloquial way of acknowledging, among other things, the unavoidable distributional shift~\cite{recht2018cifar}.} It is the job of the variational inference to make sense of this \emph{choice} and our hypothesis is that it will result in making the inference smoother, more robust to noise. More importantly, the algorithm we arrive at will be stable, in contrast to just naively formulating approximate inference for $Y$ at the opening of this section. \end{quote} 
Starting with the Gaussian kernel in Equation~\ref{eq:noisemodel}, we define the imaginary Gaussian noise model:
 \< \label{eq:decoder} p_\theta(x|z) = \mathcal{N}(x|\widehat{x}(z,\theta),\sigma^2 I_d),\>
 where the rationale for the notation $\widehat{x}(z,\theta)$ is that it is indeed the Bayes estimator of $X$ given $Z=z$: $ \widehat{x}(z,\theta)=\mathbb{E}[X|z]$. This construction can be abstracted as stated in the definition below.
\begin{definition}[Imaginary noise model] \label{def:imaginary} Consider $x$ to represent samples from $X$ and $y$ the corrupted samples by some noise/measurement process defined by $p(y|x) = \mathcal{M}(y|x,\Sigma)$, where $\Sigma$ parametrizes the noise model and  we assume a symmetric kernel: $\expectation[Y|x] = x$. The imaginary noise model for the joint density $p_\theta(x,z)=p_\theta(x|z) p_\theta(z)$ is  defined by the following:\< p_\theta(x|z) = \mathcal{M}(x|\widehat{x}(z,\theta),\Sigma), \>
where $\Sigma$ is the same set of parameters that defined the original noise model $p(y|x)$. 
\end{definition} 
The ELBO is easily derived for the imaginary Gaussian noise model as defined by Equation~\ref{eq:decoder}:
\< \label{eq:elbo} \mathcal{L}(x,\theta,\phi|\sigma) =  - \frac{1}{2\sigma^2}\mathbb{E}_{q_\phi(z|x)} \Vert x - \widehat{x}(z,\theta)\Vert^2 - \mathcal{D}_{\rm KL}[q_\phi(z|x),p_\theta(z)],\>
where $-\log(2 \pi \sigma^2)^{d/2}$ is dropped as it does not affect the optimization of the ELBO ($\sigma$ is fixed).\footnote{Not relevant for the analysis in this section, but for experiments we chose $p_\theta(z)=\mathcal{N}(z|0,I_{d_z})$, and we considered the approximate posterior to be the factorized Gaussian:~$
	q_\phi(z|x) = \prod_{i=1}^{d_z} \mathcal{N}(z_i|\mu_i(x,\phi),\sigma_i^2(x,\phi)).$} This model is named $\sigma$-VAE. We need two more ingredients before proving our main theorem. 
\clearpage
\begin{definition}[Equivalent models] \label{def:equiv}
	Consider the problem of variational inference and a fixed parametrization (a fixed architecture) for the approximate posterior $q_\phi(z|x)$ and the joint $p_\theta(x,z)$ which is used by two different models with their own sets of hyperparameters $\Sigma_1$ and $\Sigma_2$. The two models are equivalent if there exists $C_1>0$ and $C_2$ such that the following holds \< \label{eq:equiv} \mathcal{L}(x,\theta,\phi|\Sigma_2) = C_1 \mathcal{L}(x,\theta,\phi|\Sigma_1) + C_2 \>
	for all $x$ in $\mathbb{R}^d$, and all $(\theta,\phi)$ in the domains they take values in. The equivalence is denoted by:
	$$ \mathcal{L}(\Sigma_2) \equiv  \mathcal{L}(\Sigma_1)$$
\end{definition}

\begin{lemma} \label{lemma:inference} Two equivalent models learn the same set of parameters $(\theta,\phi)$ and therefore the learned models also have the same inference engine.
\begin{proof}
The proof is straightforward. It follows from Equation~\ref{eq:equiv}:
$$ \argmax_{\theta,\phi} \mathcal{L}(\theta,\phi|\Sigma_2) = \argmax_{\theta,\phi} \mathcal{L}(\theta,\phi|\Sigma_1),$$
where
$$  \mathcal{L}(\theta,\phi|\Sigma) = \frac{1}{n} \sum_{i=1}^n \mathcal{L}(x_i,\theta,\phi|\Sigma). $$
The expression ``same inference engine'' is quite intuitive: if two such learned  models are initialized with the same random seeds, they infer the same value $z$ given observation $x$ even though the two models may assign very different evidence for $x$ as measured by the ELBO.
\end{proof}	
\end{lemma}

\begin{theorem}\label{thm:sigmavae} Assume $\sigma_2$-VAE and $\sigma_1$-VAE parametrizations (architectures) are the same. Then they become equivalent via a $\beta$-VAE expansion. More precisely, $\mathcal{L}(\sigma_2) \equiv \mathcal{L}(\sigma_1, \beta)$, where $ \beta=\sigma_2^2/\sigma_1^2$.
\begin{proof}
The proof is straightforward. It follows from Equation~\ref{eq:elbo},
$$\mathcal{L}(x,\theta,\phi|\sigma_2) = \left(\frac{\sigma_1}{\sigma_2}\right)^2  \mathcal{L}(x,\theta,\phi|\sigma_1, \beta) + C, $$
where $(\sigma_1,\beta)$ is a \emph{new class} of models where $\mathcal{D}_{\rm KL}$ in Equation~\ref{eq:elbo} is multiplied by $\beta$~\cite{higgins2017beta}, where in this setup $\beta =\sigma_2^2/\sigma_1^2$. It follows from Definition~\ref{def:equiv},	
\< \mathcal{L}(\sigma_2) \equiv  \mathcal{L}(\sigma_1, \beta),\text{ where } \beta =  \left(\frac{\sigma_2}{\sigma_1}\right)^2.\>
Using Lemma~\ref{lemma:inference}, the two models $(\sigma_2)$ and $(\sigma_1,\beta)$ are the same in terms of learning and inference. Finally, as a corollary, ($\sigma$,$\beta$)-VAE is equivalent to $\sigma'$-VAE, where $\sigma'=\sigma \sqrt{\beta}$.  In other words, $\beta$ can be easily absorbed in $\sigma$.
\end{proof}
\end{theorem}
Theorem~\ref{thm:sigmavae} is just the tip of the iceberg. It can be easily extended to many other kernels that belong to \emph{exponential families}~\cite{wainwright2008graphical}. We only require them to be symmetric and their domain should also be compatible with $\mathcal{Z}$ (in this paper, $\mathcal{Z}=\mathbb{R}^{d_z}$). The ``grand theorem'' could be messy in notations due to these constraints, but to ground the ideas consider the imaginary factorized Laplace noise model:
 \< \label{eq:laplace} p_\theta(x|z) = \left(\frac{1}{2\b}\right)^d \prod_{i=1}^d \exp\left(- \frac{|x_i-\widehat{x}_i(z,\theta)|}{\b}\right),\>
where $\widehat{x}(z,\theta)=\expectation[X|z]$. It follows:
\< \label{eq:elbo2} \mathcal{L}(x,\theta,\phi|\b) =  - \frac{1}{\b}\mathbb{E}_{q_\phi(z|x)} \sum_{i=1}^d | x_i - \widehat{x}_i(z,\theta)| - \mathcal{D}_{\rm KL}[q_\phi(z|x),p_\theta(z)] ,\>
where the constant $-d \log 2\b$ is dropped ($\b$ is fixed). This model is named $\alpha$-VAE. It follows:
\begin{theorem}\label{thm:alphavae} Assume $\b_2$-VAE and $\b_1$-VAE parametrizations (architectures) are the same. Then they become equivalent via a $\beta$-VAE expansion. More precisely, \<\mathcal{L}(\b_2) \equiv \mathcal{L}(\b_1, \beta),\text{~where~}\beta =  \left(\frac{\b_2}{\b_1}\right).\>
\begin{proof} The proof is identical to the proof of Theorem~\ref{thm:sigmavae}.
\end{proof}
\end{theorem}
Next, we report some experiments we did to probe the smoothness/robustness of $\sigma$-VAE and $\b$-VAE.
\clearpage

\section{Experiments} \label{sec:experiments}
\paragraph{Network architecture.} In the experiments presented here, the encoder and decoder architectures and the learning schedule for all VAE models were the same. The posterior was the standard factorized Gaussian~\cite{kingma2013auto} and the encoder was a ConvNet with  expanding $\tt channels=(32,64,128)$, without pooling,~$\tt fc=(200)$, and the linear readout with $d_z=100$. The decoder had one hidden layer with 2000 neurons and the logistic readout.  The activation function was $u \mapsto u/(1+\exp(-u))$, a smoothed ReLU, named \emph{SiLU}~\cite{elfwing2017sigmoid} and \emph{Swish}~\cite{ramachandran2017swish}. We used the default Adam optimizer~\cite{kingma2014adam} in PyTorch~\cite{paszke2017automatic} for 100 epochs with $\tt batch size = 16$ and the constant learning rate $\tt lr=0.0001$. 

\begin{figure}[h!]
 \begin{center}
 \begin{subfigure}[$y_{ij}=x_i+\varepsilon_j,~\varepsilon_j\sim N(0,\sigma^2 I_d),~\sigma=0.9$, $x_i$ is a ``3'' from MNIST test set.]{\includegraphics[width=\textwidth]{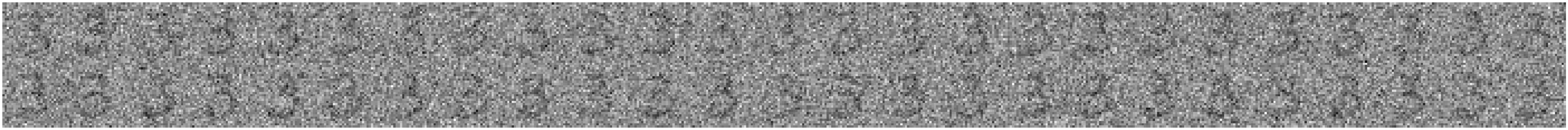}}
 \end{subfigure}
 \begin{subfigure}[$\sigma$-VAE: $\widehat{x}_\theta (z_{ij})$, where $z_{ij} \sim q_\phi(z|y_{ij})$,~$\sigma=0.9$ ]{\includegraphics[width=\textwidth]{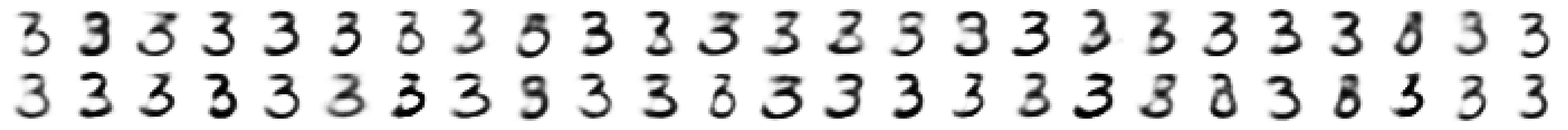}}
  \end{subfigure}	
 \begin{subfigure}[VAE: $p_\theta(z_{ij})$ from the Bernoulli decoder, where $z_{ij} \sim q_\phi(z|y_{ij})$]{\includegraphics[width=\textwidth]{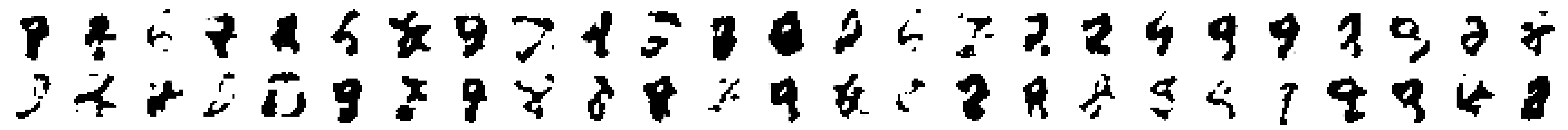}}
  \end{subfigure}   
  \begin{subfigure}[DEEN: $\widehat{x}_\vartheta(y_{ij})=y_{ij} -\sigma^2 \nabla \log f_\vartheta(y_{ij})$ where $\sigma=0.9$ (DEEN is trained with $\sigma=0.9$)]{\includegraphics[width=\textwidth]{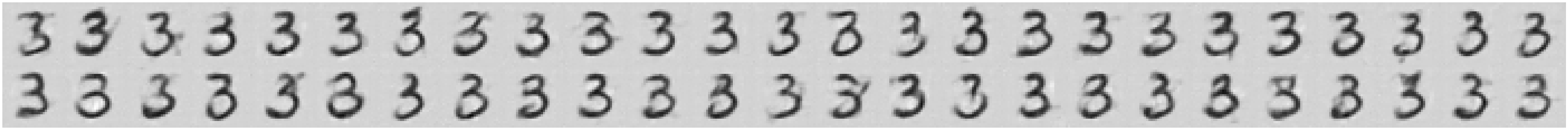}}
 \end{subfigure}
   \begin{subfigure}[$y_{ij}=x_i+\varepsilon_j,~\varepsilon_j\sim {\rm Laplace}(\b),~\alpha=0.4$, $x_i$ is a ``2'' from MNIST test set.]{\includegraphics[width=\textwidth]{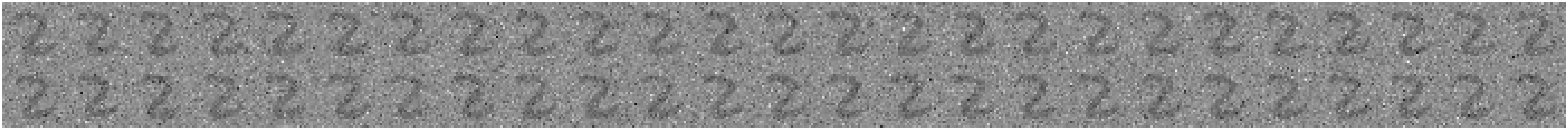}}
  \end{subfigure} 
     \begin{subfigure}[$\widehat{x}_\theta (z_{ij})$, where $z_{ij} \sim q_\phi(z|y_{ij})$ for $\alpha$-VAE trained with $\alpha=0.9$]{\includegraphics[width=\textwidth]{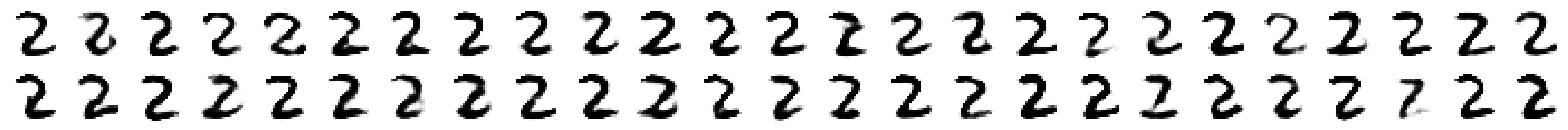}}
  \end{subfigure} 
     \begin{subfigure}[$p_\theta(z_{ij})$, where $z_{ij} \sim q_\phi(z|y_{ij})$ of the vanilla VAE, the same model as in (c)]{\includegraphics[width=\textwidth]{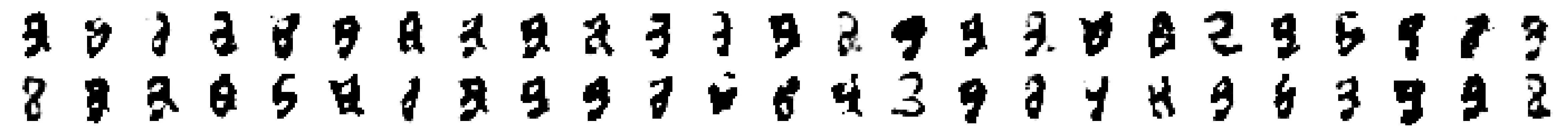}}
  \end{subfigure}   
   \begin{subfigure}[$\widehat{x}_\vartheta(y_{ij})=y_{ij} -\sigma^2 \nabla \log f_\vartheta(y_{ij})$ where $\sigma=0.6$ (DEEN is trained with $\sigma=0.6$)]{\includegraphics[width=\textwidth]{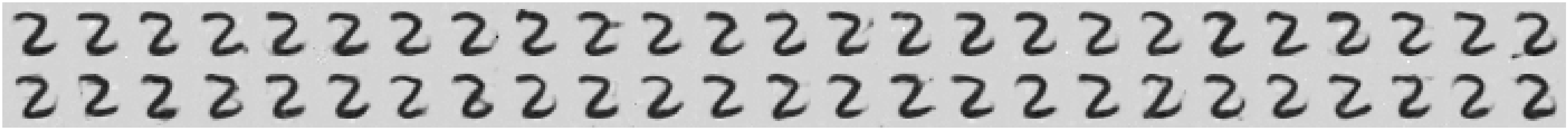}}
  \end{subfigure} 
 \caption{ {\it(testing the robustness of $\sigma$-VAE and $\b$-VAE)} {\it (a)} The samples $y_{ij}=x_i+\varepsilon_j$ are generated by adding Gaussian noise  to a sample $x_i$ from the MNIST test set. Here 50 such noisy samples are shown. {\it (b)} The noisy samples are given to $\sigma$-VAE which is trained seeing only clean samples. Here are the results for $\widehat{x}(z_{ij})$ of the decoder obtained after inferring the $z_{ij}$ by the encoder. {\it (c)} The same noisy samples are given to vanilla VAE with the same architecture and trained with the same learning schedule as $\sigma$-VAE. The VAE simply breaks down in this regime. {\it (d)} The same noisy samples are given to DEEN after learning the energy function $f_\vartheta(\cdot)$ for $\sigma=0.9$. The Bayes estimator of $X$ are shown here. Note that this is a deterministic computation: given the noisy data from (a) DEEN always return the same answer. Also, visually (this is subjective), $\sigma$-VAE shows more ``understanding'' of the handwritten digits in interpreting noise as the denoised samples have different styles as opposed to DEEN's mechanical (but powerful) denoising computation. {\it (e)} The experiments were repeated by training $\alpha$-VAE as defined by Equations~\ref{eq:laplace} and~\ref{eq:elbo2}. The additive noise is sampled from the factorized Laplace distribution in $\mathbb{R}^d$, denoted by  Laplace($\b$), parametrized by $\alpha=0.4$. {\it (f)} The $\alpha$-VAE trained with $\alpha=0.9$ is tested. To reiterate, noisy samples are never seen in training $\alpha$-VAE. {\it (g)} As before, the vanilla VAE simply breaks down. {\it (h)}  For DEEN, the denoising results were poor when we tested the model on $f_\vartheta(0.9)$, but $f_\vartheta(0.6)$ does a good job, but again its inference-free mechanical nature is fully visible (see Remark 8 in~\cite{saremi2019neural} regarding the ``gray background''). For more examples see Appendix~\ref{sec:appendix:laplace} and~\ref{sec:appendix:vizsigma} where we also include experiments on high levels of salt-and-pepper noise. }
 \label{fig:xhat}
 \end{center}
 \end{figure}

{\bf Probing the smoothness/robustness of $\sigma$-VAE and $\alpha$-VAE.} Imaginary noise models were in part motivated by making the variational inference in VAEs more robust. This was tested by showing images with large amounts of noise to the trained models. The first set of experiments are presented in Figure~\ref{fig:xhat} (a-d) for the Gaussian noise with the maximum noise level $\sigma=0.9$ which we ran our experiments. For a geometric understanding of this noise level we refer to~\cite{saremi2019neural}, but the noise also happens to be quite high for \emph{our} visual system.  As it is clear, vanilla VAE simply breaks down, but quite remarkably the $\sigma$-VAE makes reasonable inferences even though it has not seen any noisy data during training. The $\sigma$-VAE is also compared with DEEN. Note that DEEN's learning objective (Eq.~\ref{eq:deen}) is essentially a denoising objective by learning the energy function parametrized by a ConvNet where its gradient $\nabla f_\vartheta$ is used such that expected squared deviation of the Bayes estimator from the clean data is minimized: in short, DEEN is a denoising powerhouse. On the other hand, the learning objective in $\sigma$-VAE is not based on denoising\textemdash \emph{it does not see noisy samples}\textemdash and it is a surprising result that the model does not break down for such large amounts of noise. The experiments were repeated for $\alpha$-VAE as defined by Eqs.~\ref{eq:laplace} and~\ref{eq:elbo2} and a set of them is presented in Fig.~\ref{fig:xhat} (e-h).

{\bf Scaling laws.} Quantitatively, we observed a power law that the $\sigma$-VAE seems to obey given by $\mathcal{D}_{\rm KL} \sim \sigma^{-\nu}$. The values $\langle \mathcal{D}_{\rm KL}[q(z|x),p(z)] \rangle$ for trained $\sigma$-VAE models are reported in Table~\ref{table:powerlaw} where the exponent $\nu \approx \it 1.15$ was fit by linear regression on the log-log plot. We also report the mean squared error $\mathbb{E}_{q_\phi(z|x)} \Vert x-\widehat{x}(z) \Vert^2$  in Table~\ref{table:mse}. Putting the results from the two tables together, we observe $\mathcal{D}_{\rm KL}/\mathcal{L} = - 0.481 \pm 0.045,$ fairly constant in $[0.1,0.9]$. Both results generalize to $\alpha$-VAE: $\mathcal{D}_{\rm KL} \sim \b^{-\nu}$  with the exponent $\nu\approx \it 0.636$ and  $\mathcal{D}_{\rm KL}/\mathcal{L} = -0.368\pm  0.005$ (see Appendix~\ref{sec:appendix:laplace}). It follows from Theorem~\ref{thm:sigmavae} that $\mathcal{D}_{\rm KL}\sim \beta^{-\gamma}$, where $\gamma=\nu/2$ for $\sigma$-VAE (and $\gamma=\nu$ for $\alpha$-VAE). {\it To our knowledge, such scaling laws have never been reported in the literature on variational inference.}

{\bf A quantitative evaluation of the robustness of $\sigma$-VAE.} Here, we report some experiments on using $\sigma$-VAE for provable robust classification with \emph{randomized smoothing}. A survey of randomized smoothing~\cite{cohen2019certified} is beyond the scope of this work, but the idea is to construct smoothed classifiers based on $\widehat{x}(z,\theta)$ as was done in~\cite{saremi2020provable} using DEEN.  We ran experiments using the machinery of XHAT in~\cite{saremi2020provable} by replacing $\widehat{x}_\vartheta(y)=y -\sigma^2 \nabla f_\vartheta(y)$ with $\widehat{x}_\theta(z),~ z\sim q_\phi(z|y)$. Experiments were performed for $\sigma=0.7$, and XHAT (using DEEN) came up on top with a 35+\% margin over the range of radii $[0.5,1.5]$.  Closing this gap is a good challenge! Note that DEEN is especially well suited here due to its ``direct'' denoising machinery from $Y$ to $X$. We believe an approach based on smoothed variational inference is more appealing conceptually, but one may need to explore a much higher dimensional $Z$.

\begin{remark}
	Regarding DEEN's denoising performance for the Laplace noise (Figure~\ref{fig:xhat}h), one can in principle replace~(\ref{eq:xhat}) and develop a new learning objective using the empirical Bayes estimator for the Laplace measurement kernel, but the algorithm is not tractable due to the integrals involved. On the other hand, $\alpha$-VAE is quite simple to implement as defined by its decoder~(\ref{eq:laplace}) and the ELBO~(\ref{eq:elbo2}).
\end{remark}

\begin{table}[h!]
\begin{center}
\begin{tabular}{c| c c c c c c c c c }
\toprule
$\sigma$   & $\it 0.1$ & $\it 0.2$& $\it 0.3$& $\it 0.4$ & $\it 0.5$ & $\it 0.6$ & $\it 0.7$ & $\it 0.8$ & $\it 0.9$ \\
\midrule 
$\mathcal{D}_{\rm KL}$ & 108 & 56.6 	& 37.2 & 27.6 & 20.5 & 16.4 & 12.6 & 10.4 & 8.61 \\
\bottomrule
\end{tabular}
\end{center}
\caption{ The $\sigma$-VAE on MNIST obeys the power law $\mathcal{D}_{\rm KL} \sim \sigma^{-\nu}$, $\nu \approx 1.15$ } 
\label{table:powerlaw}
\end{table}
\vspace{-0.09cm}
\begin{table}[h!]
\begin{center}
\begin{tabular}{c| c c c c c c c c c }
\toprule
$\sigma$ &  $\it 0.1$ & $\it 0.2$& $\it 0.3$& $\it 0.4$ & $\it 0.5$ & $\it 0.6$ & $\it 0.7$ & $\it 0.8$ & $\it 0.9$ \\
\midrule 
$\langle \mathbb{E}_{q_{\phi}(z|x)} \Vert x-\widehat{x}(z)  \Vert^2 \rangle$ &  1.97 & 4.01	& 6.21 & 8.17 & 10.6 & 13.0 & 15.6 & 18.1 & 20.7 \\
\bottomrule
\end{tabular}
\end{center}
\caption{ The reconstruction evaluation of the $\sigma$-VAE as measured by the mean-squared error.}  
\label{table:mse}
\end{table}

\clearpage
\begin{remark}[on the KL scaling laws] In nature, power laws fall into universality classes~\cite{wilson1979problems}. Here, we do not have a clear answer on whether there is any notion of ``universality'' regarding the empirical observations $\mathcal{D}_{\rm KL} \sim \sigma^{-\nu}$ and $\mathcal{D}_{\rm KL} \sim \alpha^{-\nu}$. Formalizing the XYZ hypothesis (see Fig.~\ref{fig:XYZ} and Sec.~\ref{sec:discussion}) could be the key to this problem as both $\sigma$ and $\alpha$ set the scale on how smooth $Y$ is.
\end{remark}

\section{Related Work} \label{sec:related}
The genesis of this work was to add \emph{inference} to neural empirical Bayes~\cite{saremi2019neural} which had been developed for learning \emph{smoothed densities}. But we ended up approaching the problem backwards by making variational autoencoders~\cite{kingma2013auto,rezende2014stochastic} more smooth. We were also motivated by~\cite{saremi2020provable} and how the smoothed densities learned by DEEN was integrated into the framework of randomized smoothing~\cite{cohen2019certified} for certified robust classification. It is straightforward to set up $\sigma$-VAE (or better-designed smoothed variational autoencoders) for certified robust classification, but as we reported there are technical challenges to compete with DEEN since $\sigma$-VAE is not designed around denoising.

The connections to $\beta$-VAE  are indeed surprising since  this work was motivated by very different set of problems than learning disentangled representations~\cite{higgins2017beta,chen2018isolating, burgess2018understanding}. The fact that in imaginary Gaussian noise models a $\beta$-VAE can be easily \emph{mapped} to a ``proper'' $\beta=1$ model together with its immediate generalization to the Laplace distribution is a hint that there must be richer structures than what we have explored here. Along these lines, we should also mention~\cite{alemi2016deep} for  another perspective on the $\beta$. From our perspective, we are eventually interested in (very) high-dimensional $Z$ and \emph{highly distributed} representations where one has to rethink the topic of disentanglement from ground up. 

Regarding Gaussian decoders, the most comprehensive study we could find was~\cite{dai2019diagnosing} with motivations to improve the sample quality of VAEs. We were not interested in that problem in this paper. Regarding the fixed $\sigma$, we should mention~\cite{bousquet2017optimal, doersch2016tutorial} for more discussions that also goes around sample quality, where in fact it appears $\sigma=1.0$ had already been looked at, but only for generative modeling. In this work, we stopped at $\sigma=0.9$ (see Appendix~\ref{sec:appendix:samples} for samples generated by $z \mapsto \widehat{x}(z,\theta)$). 

The topic of high $d_z$ is discussed in~\cite{dai2019diagnosing} for VAEs, and it also had a precedence in~\cite{mehrjou2017annealed} where they set $d_z =d$ to address the problem of mode collapse in GANs~\cite{goodfellow2014generative}. For the problem of \emph{smoothed} variational inference,  we believe high $d_z$ is not really a choice but a requirement, but unfortunately, we do not have a clear answer on how to compute the threshold for ``high''. Ideally, we would like to explore the regime of large $\sigma$ and $d_z\gg d$, however that comes with its own computational challenges.


\section{Discussion} \label{sec:discussion}
\emph{We think $Y$ is somehow represented in $\mathcal{Z}$ even though the model only sees clean samples from $X$.} This is the XYZ hypothesis. The hypothesis is mainly based on intuition, but it is supported by the fact that the $\sigma$-VAE was robust to high levels of noise.  It is also supported by the power law $\mathcal{D}_{\rm KL}\sim \sigma^{-\nu}$,  which means geometrically that the representation of the posterior gets expanded for larger $\sigma$. This is  expected from the hypothesis since the manifold of $Y$ is also expanded/smoothed compared to $X$. This smoothed representation is well suited for robust inference as one would intuitively expect.~$\bullet$~Robust variational inference on MNIST is not solved! But regarding scaling $\sigma$-VAE to more complex datasets, the computational challenge is that a large latent space might be necessary. However, that may also come with nice properties, e.g. we may not need to worry as much about the choice of the factorized Gaussian for the posterior. This is related to our discussion of the \emph{fundamental duality} in latent variable models and the fact that $Y$ has a more tractable distribution than $X$.~$\bullet$~The fact that in imaginary noise models there are equivalence classes in terms of learning and inference is exciting but also alarming. The $\sigma$-VAE is indeed grounded in the foundations of variational inference\textemdash that was the starting point and we went at great length to break the inference, but we failed! However, what meaning is left to the evidence lower bound when two \emph{equivalent} models can be easily constructed with different ELBOs as we saw in the proof of Theorem~\ref{thm:sigmavae}? Although this was not the intention, but this paper also aims at framing the topic of evaluations in VAEs around learning representations that are suited for robust inference.~$\bullet$~Finally, imaginary noise models are indeed quite philosophical  in their utterly unrealistic and pessimistic view of the world. But at the computational level, this study is yet another demonstration of the power of  learning and inference in latent variable models.  Here, the inference engine of the $\sigma$-VAE manages to take the imaginary noise model into consideration and learns a representation that is smoother $\approx$ more robust.

\section*{Broader Impact}
This work aimed at formulating a notion of \emph{smoothed variational inference} and we also discovered formal links between the implicit smoothing achieved in \emph{imaginary noise models} and learning disentangled representations that motivated the development of $\beta$-VAE, encapsulated in Theorems~\ref{thm:sigmavae} and~\ref{thm:alphavae}. In this paper, we probed the smoothing quantitatively with the Kullback-Leibler divergence \emph{power laws} and qualitatively with robustness to  large amounts of noise, but we were interested in that problem in itself: in the total absence of adversaries \emph{per se}. It remains to be seen whether the implicit smoothing with imaginary noise can be as effective as the explicit smoothing with real noise. Ultimately, learning must be viewed as \emph{synonymous} with robust learning and inference synonymous with robust inference, otherwise we can never fully trust a machine learning system in applications. In addition, this ``synonymity'' is also aligned with the goal of engineering systems whose level of intelligence, although indeed extremely limited, at least could have a ``flavor'' akin to \emph{our} intelligence.

\section*{Acknowledgments} I would like to thank Christian Osendorfer for discussions, and for his  valuable comments on the manuscript. I am also grateful to Rupesh Srivastava and Giorgio Giannone for discussions.

\bibliographystyle{plain}

\bibliography{svae}

\begin{thebibliography}{10}

\bibitem{alemi2016deep}
Alexander~A Alemi, Ian Fischer, Joshua~V Dillon, and Kevin Murphy.
\newblock Deep variational information bottleneck.
\newblock {\em arXiv preprint arXiv:1612.00410}, 2016.

\bibitem{anderson1987mean}
James~R Anderson and Carsten Peterson.
\newblock A mean field theory learning algorithm for neural networks.
\newblock {\em Complex Systems}, 1:995--1019, 1987.

\bibitem{bengio2013representation}
Yoshua Bengio, Aaron Courville, and Pascal Vincent.
\newblock Representation learning: A review and new perspectives.
\newblock {\em IEEE Transactions on Pattern Analysis and Machine Intelligence},
  35(8):1798--1828, 2013.

\bibitem{bousquet2017optimal}
Olivier Bousquet, Sylvain Gelly, Ilya Tolstikhin, Carl-Johann Simon-Gabriel,
  and Bernhard Schoelkopf.
\newblock From optimal transport to generative modeling: the vegan cookbook.
\newblock {\em arXiv preprint arXiv:1705.07642}, 2017.

\bibitem{burgess2018understanding}
Christopher~P Burgess, Irina Higgins, Arka Pal, Loic Matthey, Nick Watters,
  Guillaume Desjardins, and Alexander Lerchner.
\newblock Understanding disentangling in $\beta$-{VAE}.
\newblock {\em arXiv preprint arXiv:1804.03599}, 2018.

\bibitem{chen2018isolating}
Tian~Qi Chen, Xuechen Li, Roger~B Grosse, and David~K Duvenaud.
\newblock Isolating sources of disentanglement in variational autoencoders.
\newblock In {\em Advances in Neural Information Processing Systems}, pages
  2610--2620, 2018.

\bibitem{cohen2019certified}
Jeremy~M Cohen, Elan Rosenfeld, and J~Zico Kolter.
\newblock Certified adversarial robustness via randomized smoothing.
\newblock {\em arXiv preprint arXiv:1902.02918}, 2019.

\bibitem{dai2019diagnosing}
Bin Dai and David Wipf.
\newblock Diagnosing and enhancing {VAE} models.
\newblock {\em arXiv preprint arXiv:1903.05789}, 2019.

\bibitem{doersch2016tutorial}
Carl Doersch.
\newblock Tutorial on variational autoencoders.
\newblock {\em arXiv preprint arXiv:1606.05908}, 2016.

\bibitem{elfwing2017sigmoid}
Stefan Elfwing, Eiji Uchibe, and Kenji Doya.
\newblock Sigmoid-weighted linear units for neural network function
  approximation in reinforcement learning.
\newblock {\em arXiv preprint arXiv:1702.03118}, 2017.

\bibitem{goodfellow2014generative}
Ian Goodfellow, Jean Pouget-Abadie, Mehdi Mirza, Bing Xu, David Warde-Farley,
  Sherjil Ozair, Aaron Courville, and Yoshua Bengio.
\newblock Generative adversarial nets.
\newblock In {\em Advances in neural information processing systems}, pages
  2672--2680, 2014.

\bibitem{higgins2017beta}
Irina Higgins, Loic Matthey, Arka Pal, Christopher Burgess, Xavier Glorot,
  Matthew Botvinick, Shakir Mohamed, and Alexander Lerchner.
\newblock $\beta$-{VAE}: Learning basic visual concepts with a constrained
  variational framework.
\newblock In {\em International Conference on Learning Representations}, 2017.

\bibitem{hinton1986learning}
Geoffrey~E Hinton and Terrence~J Sejnowski.
\newblock Learning and relearning in {B}oltzmann machines.
\newblock {\em Parallel Distributed Processing: Explorations in the
  Microstructure of Cognition}, 1(282-317):2, 1986.

\bibitem{hyvarinen2005estimation}
Aapo Hyv{\"a}rinen.
\newblock Estimation of non-normalized statistical models by score matching.
\newblock {\em Journal of Machine Learning Research}, 6(Apr):695--709, 2005.

\bibitem{im2017denoising}
Daniel Im~Jiwoong Im, Sungjin Ahn, Roland Memisevic, and Yoshua Bengio.
\newblock Denoising criterion for variational auto-encoding framework.
\newblock In {\em Thirty-First AAAI Conference on Artificial Intelligence},
  2017.

\bibitem{jordan1999introduction}
Michael~I Jordan, Zoubin Ghahramani, Tommi~S Jaakkola, and Lawrence~K Saul.
\newblock An introduction to variational methods for graphical models.
\newblock {\em Machine learning}, 37(2):183--233, 1999.

\bibitem{kingma2014adam}
Diederik~P Kingma and Jimmy Ba.
\newblock Adam: A method for stochastic optimization.
\newblock {\em arXiv preprint arXiv:1412.6980}, 2014.

\bibitem{kingma2013auto}
Diederik~P Kingma and Max Welling.
\newblock Auto-encoding variational {B}ayes.
\newblock {\em arXiv preprint arXiv:1312.6114}, 2013.

\bibitem{lecun1998gradient}
Yann LeCun, L{\'e}on Bottou, Yoshua Bengio, and Patrick Haffner.
\newblock Gradient-based learning applied to document recognition.
\newblock {\em Proceedings of the IEEE}, 86(11):2278--2324, 1998.

\bibitem{ledoux2001concentration}
Michel Ledoux.
\newblock {\em The concentration of measure phenomenon}.
\newblock Number~89. American Mathematical Soc., 2001.

\bibitem{mackay2003information}
David~JC MacKay.
\newblock {\em Information theory, inference and learning algorithms}.
\newblock Cambridge University Press, 2003.

\bibitem{mehrjou2017annealed}
Arash Mehrjou, Bernhard Sch{\"o}lkopf, and Saeed Saremi.
\newblock Annealed generative adversarial networks.
\newblock {\em arXiv preprint arXiv:1705.07505}, 2017.

\bibitem{miyasawa1961empirical}
Koichi Miyasawa.
\newblock An empirical {B}ayes estimator of the mean of a normal population.
\newblock {\em Bulletin of the International Statistical Institute},
  38(4):181--188, 1961.

\bibitem{paszke2017automatic}
Adam Paszke, Sam Gross, Soumith Chintala, Gregory Chanan, Edward Yang, Zachary
  DeVito, Zeming Lin, Alban Desmaison, Luca Antiga, and Adam Lerer.
\newblock Automatic differentiation in {PyTorch}.
\newblock 2017.

\bibitem{ramachandran2017swish}
Prajit Ramachandran, Barret Zoph, and Quoc~V Le.
\newblock Swish: a self-gated activation function.
\newblock {\em arXiv preprint arXiv:1710.05941}, 7, 2017.

\bibitem{raphan2011least}
Martin Raphan and Eero~P Simoncelli.
\newblock Least squares estimation without priors or supervision.
\newblock {\em Neural computation}, 23(2):374--420, 2011.

\bibitem{recht2018cifar}
Benjamin Recht, Rebecca Roelofs, Ludwig Schmidt, and Vaishaal Shankar.
\newblock Do {CIFAR}-10 classifiers generalize to {CIFAR}-10?
\newblock {\em arXiv preprint arXiv:1806.00451}, 2018.

\bibitem{rezende2014stochastic}
Danilo~Jimenez Rezende, Shakir Mohamed, and Daan Wierstra.
\newblock Stochastic backpropagation and approximate inference in deep
  generative models.
\newblock {\em arXiv preprint arXiv:1401.4082}, 2014.

\bibitem{robbins1956empirical}
Herbert Robbins.
\newblock An empirical {B}ayes approach to statistics.
\newblock In {\em Proc. Third Berkeley Symp.}, volume~1, pages 157--163, 1956.

\bibitem{robbins2012selected}
Herbert Robbins.
\newblock {\em Selected papers}.
\newblock Springer, 2012.

\bibitem{roweis2000nonlinear}
Sam~T Roweis and Lawrence~K Saul.
\newblock Nonlinear dimensionality reduction by locally linear embedding.
\newblock {\em Science}, 290(5500):2323--2326, 2000.

\bibitem{saremi2019approximating}
Saeed Saremi.
\newblock On approximating $\nabla f $ with neural networks.
\newblock {\em arXiv preprint arXiv:1910.12744}, 2019.

\bibitem{saremi2019neural}
Saeed Saremi and Aapo Hyv{{\"a}}rinen.
\newblock Neural empirical {B}ayes.
\newblock {\em Journal of Machine Learning Research}, 20(181):1--23, 2019.

\bibitem{saremi2018deep}
Saeed Saremi, Arash Mehrjou, Bernhard Sch{\"o}lkopf, and Aapo Hyv{\"a}rinen.
\newblock Deep energy estimator networks.
\newblock {\em arXiv preprint arXiv:1805.08306}, 2018.

\bibitem{saremi2020provable}
Saeed Saremi and Rupesh Srivastava.
\newblock Provable robust classification via learned smoothed densities.
\newblock {\em arXiv preprint arXiv:2005.04504}, 2020.

\bibitem{saul2003think}
Lawrence~K Saul and Sam~T Roweis.
\newblock Think globally, fit locally: unsupervised learning of low dimensional
  manifolds.
\newblock {\em Journal of Machine Learning Research}, 4(Jun):119--155, 2003.

\bibitem{tao2012topics}
Terence Tao.
\newblock {\em Topics in random matrix theory}.
\newblock American Mathematical Society, 2012.

\bibitem{tenenbaum2000global}
Joshua~B Tenenbaum, Vin De~Silva, and John~C Langford.
\newblock A global geometric framework for nonlinear dimensionality reduction.
\newblock {\em Science}, 290(5500):2319--2323, 2000.

\bibitem{tenenbaum2011grow}
Joshua~B Tenenbaum, Charles Kemp, Thomas~L Griffiths, and Noah~D Goodman.
\newblock How to grow a mind: statistics, structure, and abstraction.
\newblock {\em Science}, 331(6022):1279--1285, 2011.

\bibitem{wainwright2008graphical}
Martin~J Wainwright and Michael~I Jordan.
\newblock Graphical models, exponential families, and variational inference.
\newblock {\em Foundations and Trends in Machine Learning}, 1(1--2):1--305,
  2008.

\bibitem{wilson1979problems}
Kenneth~G Wilson.
\newblock Problems in physics with many scales of length.
\newblock {\em Scientific American}, 241(2):158--179, 1979.

\end{thebibliography}

\clearpage

\appendix

  \section{$\alpha$-VAE: smoothed variational inference via imaginary Laplace noise} \label{sec:appendix:laplace}
  
\begin{figure}[h!]
 \begin{center} 
   \begin{subfigure}[$x_i$ from the MNIST test set]{\includegraphics[width=\textwidth]{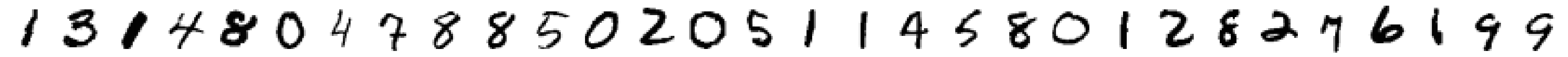}}
  \end{subfigure} 
     \begin{subfigure}[ $y_{ij}=x_i+\varepsilon_j,~\varepsilon_j\sim {\rm Laplace}(\b)$, where $\b=0.4$ ]{\includegraphics[width=\textwidth]{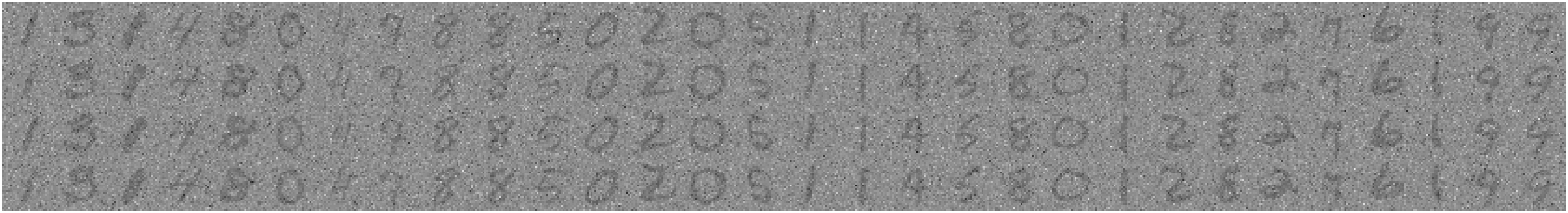}}
  \end{subfigure} 
     \begin{subfigure}[$\widehat{x}_\theta(z_{ij})$, where $z_{ij} \sim q_\phi(z|y_{ij})$ for $\b$-VAE trained with $\b=0.9$ ]{\includegraphics[width=\textwidth]{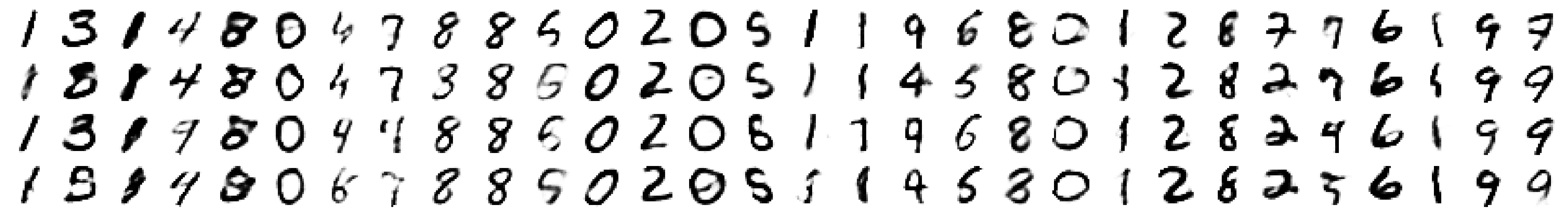}}
  \end{subfigure}   
   \begin{subfigure}[$\widehat{x}_\theta(y_{ij})=y_{ij} -\sigma^2 \nabla f_\vartheta(y_{ij})$ for DEEN trained with $\sigma=0.6$]{\includegraphics[width=\textwidth]{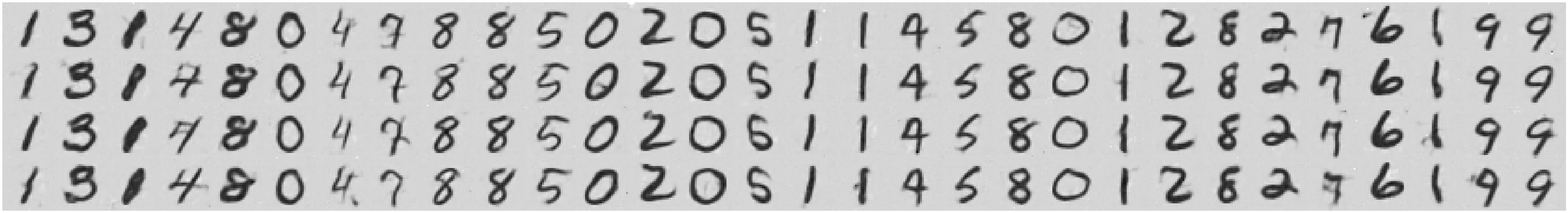}}
  \end{subfigure} 
   \begin{subfigure}[$p_\theta(z_{ij})$ of the Bernoulli decoder of the vanilla VAE, where $z_{ij} \sim q_\phi(z|y_{ij})$]{\includegraphics[width=\textwidth]{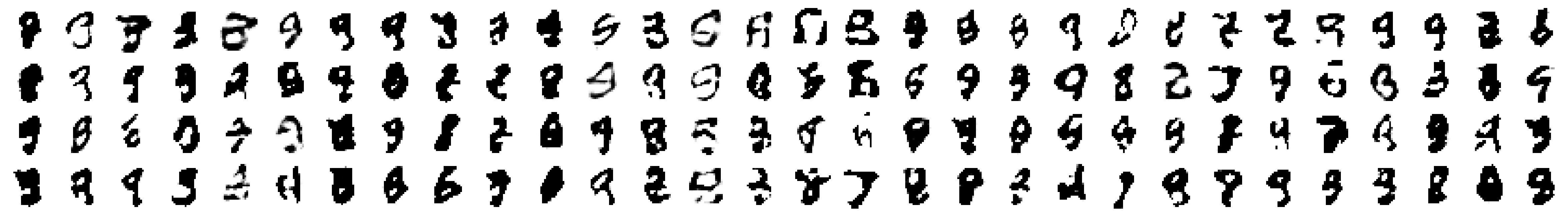}}
  \end{subfigure}   
 \caption{{\it (Robustness of $\b$-VAE to Laplace noise)} See Figure~\ref{fig:xhat} in the paper for captions.}
 \label{fig:laplace}
 \end{center}
 \end{figure}

We repeated the experiments reported in Section~\ref{sec:experiments} for the imaginary Laplace noise model as defined by Eqs.~\ref{eq:laplace} and~\ref{eq:elbo2}. The model is named $\b$-VAE. In Table~\ref{table:appendix:laplace}, we report $\mathcal{D}_{\rm KL} \sim \b^{-\nu}$ (short for $\langle \mathcal{D}_{\rm KL}[q_\phi(z|x),p_\theta(z)]\rangle$ for the learned models where $\langle \cdot \rangle$ is the expectation over the test set) and the exponent $\nu \approx  0.636$ was fit with linear regression $\log \mathcal{D}_{\rm KL} = -\nu \log \b + C$ with the p-value = $5.3\times 10^{-12}$. In Table~\ref{table:appendix:laplace:ratios} we report the ratios we discussed in Section~\ref{sec:experiments}, where again they remain approximately constant in the range $[0.1,0.9]$. In Fig.~\ref{fig:laplace}, we report the robustness of $\b$-VAE to noise.

 \begin{table}[h!]
\begin{center}
\begin{tabular}{c| c c c c c c c c c }
\toprule
$\b$   & $\it 0.1$ & $\it 0.2$& $\it 0.3$& $\it 0.4$ & $\it 0.5$ & $\it 0.6$ & $\it 0.7$ & $\it 0.8$ & $\it 0.9$ \\
\midrule 
$\mathcal{D}_{\rm KL}$ & 78.6 &  52.3 	& 40.3 & 33.3 & 28.9 & 25.7 & 23.5 & 21.3 & 19.4 \\
\bottomrule
\end{tabular}
\end{center}
\caption{ The $\b$-VAE on MNIST obeys the power law $\mathcal{D}_{\rm KL} \sim \b^{-\nu}$, $\nu \approx 0.636$} 
\label{table:appendix:laplace}
\end{table}
 \begin{table}[h!]
\begin{center}
\begin{tabular}{c| c c c c c c c c c }
\toprule
$\b$   & $\it 0.1$ & $\it 0.2$& $\it 0.3$& $\it 0.4$ & $\it 0.5$ & $\it 0.6$ & $\it 0.7$ & $\it 0.8$ & $\it 0.9$ \\
\midrule 
$-\mathcal{D}_{\rm KL}/\mathcal{L}(\sigma)$ & 0.36 &  0.37 	& 0.37& 0.37 & 0.37 & 0.37 & 0.37 & 0.37 & 0.36 \\
\bottomrule
\end{tabular}
\end{center}
\caption{ The ratio $\mathcal{D}_{\rm KL}/\mathcal{L}=-0.368\pm  0.005$. Note that $-d \log 2\b$ is dropped from the ELBO  as discussed in Section~\ref{sec:SVAE} and not included here either.} 
\label{table:appendix:laplace:ratios}
\end{table}
 \clearpage

\section{More visualizations on the robustness of $\sigma$-VAE} \label{sec:appendix:vizsigma}
 \begin{figure}[h!]
 \begin{center} 
   \begin{subfigure}[$x_i$ from the MNIST test set]{\includegraphics[width=\textwidth]{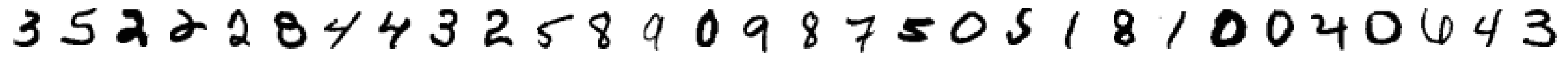}}
  \end{subfigure} 
     \begin{subfigure}[$y_{ij}=x_i+\varepsilon_j,~\varepsilon_j\sim N(0,\sigma^2 I_d)$, where $\sigma=0.9$ ]{\includegraphics[width=\textwidth]{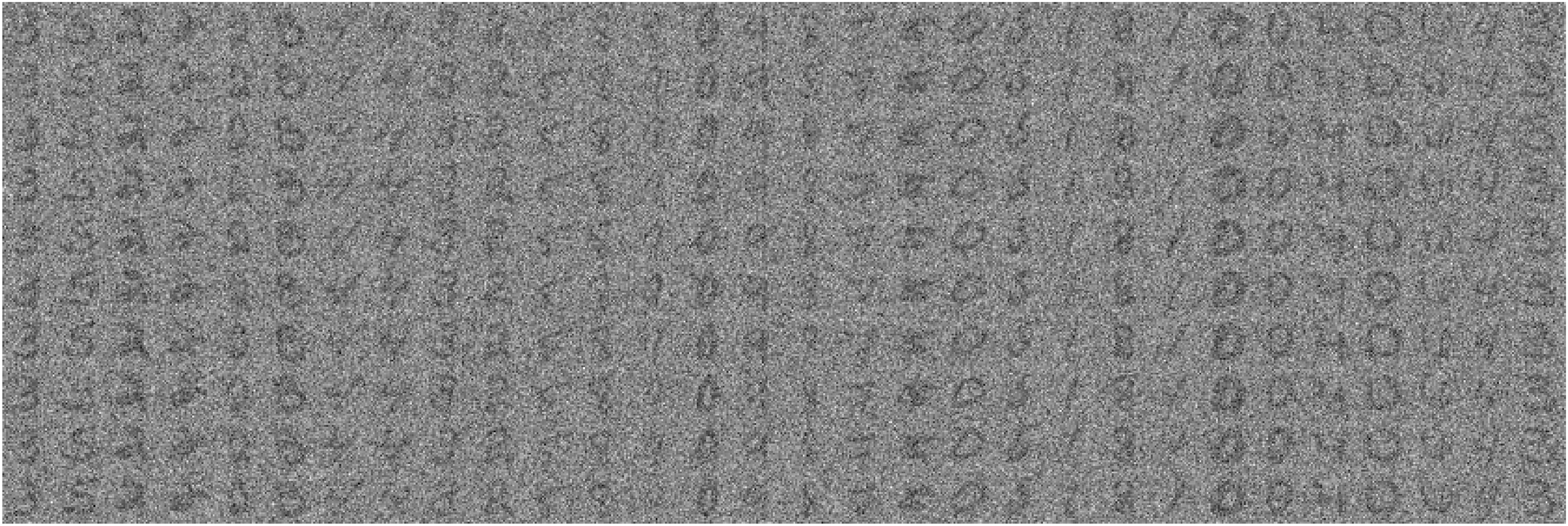}}
  \end{subfigure} 
     \begin{subfigure}[$\widehat{x}_\theta(z_{ij})$, where $z_{ij} \sim q_\phi(z|y_{ij})$ for $\sigma$-VAE with $\sigma=0.9$ ]{\includegraphics[width=\textwidth]{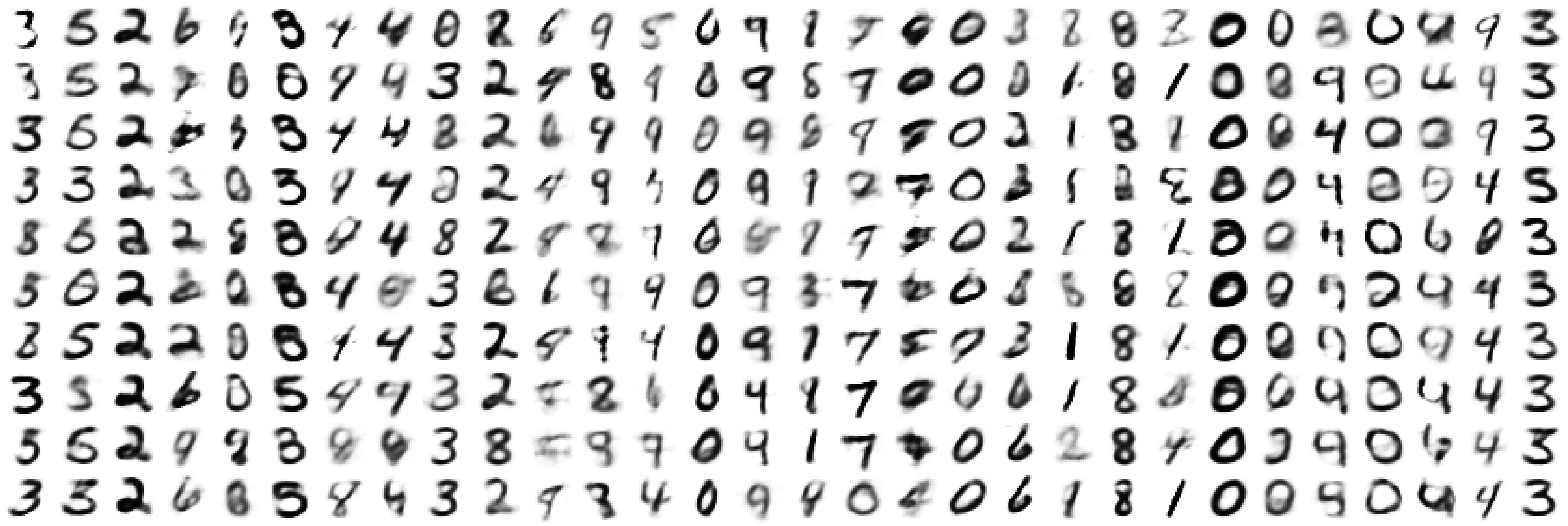}}
  \end{subfigure}   
   \begin{subfigure}[$\widehat{x}_\theta(y_{ij})=y_{ij} -\sigma^2 \nabla f_\vartheta(y_{ij})$ for DEEN trained with $\sigma=0.9$]{\includegraphics[width=\textwidth]{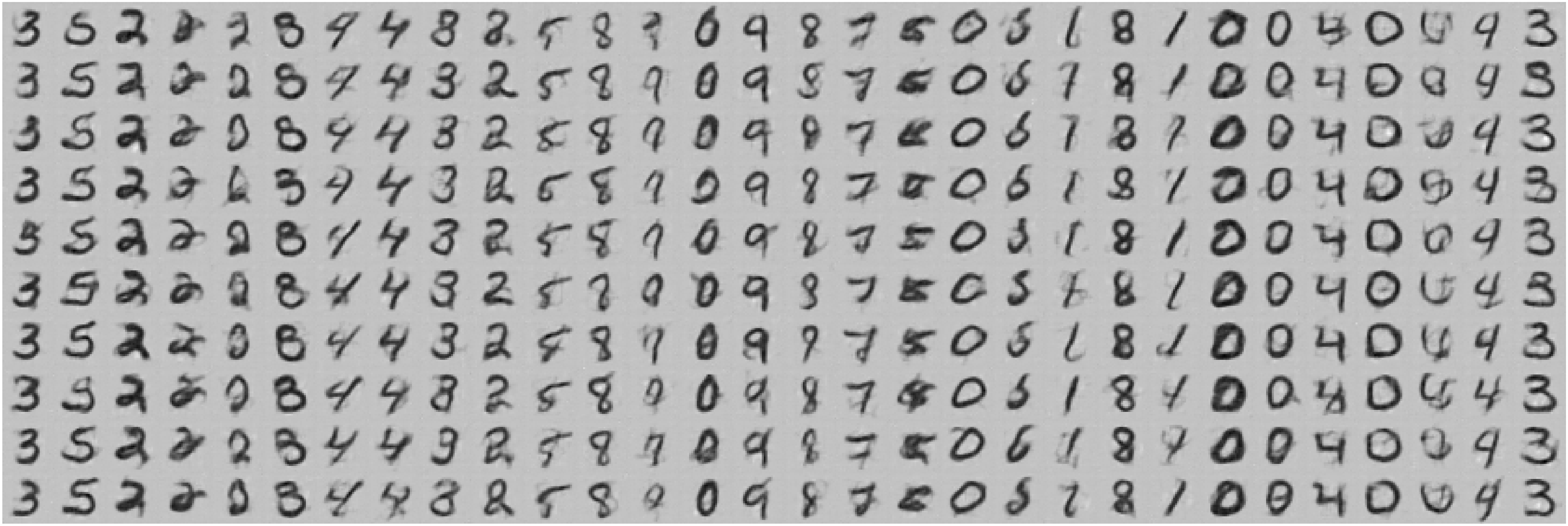}}
  \end{subfigure} 
 \caption{ (More examples on robustness of the $\sigma$-VAE to Gaussian noise.) (a) The samples $x_i$ from MNIST. (b) The samples $y_{ij}=x_i+\varepsilon_j,~\varepsilon_j\sim N(0,\sigma^2 I_d)$, where $\sigma=0.9$. These noisy samples are held fixed for the rest. (c) The Bayes estimator $\widehat{x}_\theta(z_{ij})=\expectation[X|z_{ij}]$ is shown here, where $z_{ij} \sim q_\phi(z|y_{ij})$. The $\sigma$-VAE is trained with $\sigma=0.9$. Note that $\sigma$-VAE does not see any noisy samples during training. (d) The Bayes estimator $\widehat{x}_\vartheta(y_{ij})=y_{ij} -\sigma^2 \nabla f_\vartheta(y_{ij})$, where $f_\vartheta$ is learned with $\sigma=0.9$. DEEN is designed around least-squares denoising but one can spot examples where $\sigma$-VAE's  ``thought process'' is in display in interpreting the noise in comparison with DEEN's inference-free deterministic computation (see Remark 8 in~\cite{saremi2019neural} regarding the gray background). The vanilla VAE results are not shown as the model simply breaks down (see Figure~\ref{fig:xhat} for examples).}
 \label{fig:axhat}
 \end{center}
 \end{figure}
 \clearpage
 
Below we compare the robustness to noise of $\sigma$-VAEs trained with different $\sigma$ in $[0.1,0.9]$. From the power law $\mathcal{D}_{\rm KL} \sim \sigma^{-\nu}$ ($\nu \approx 1.15$) we expect the posterior to be more expanded and the model to be more robust to noise for larger $\sigma$, and this is also consistent with the XYZ hypothesis.
  \begin{figure}[b!]
 \begin{center} 
   \begin{subfigure}[$y_{ij}=x_i+\varepsilon_j,~\varepsilon_j\sim N(0,\sigma^2 I_d),~\sigma=0.9$, $x_i$ is a ``5'' from MNIST test set]{\includegraphics[width=\textwidth]{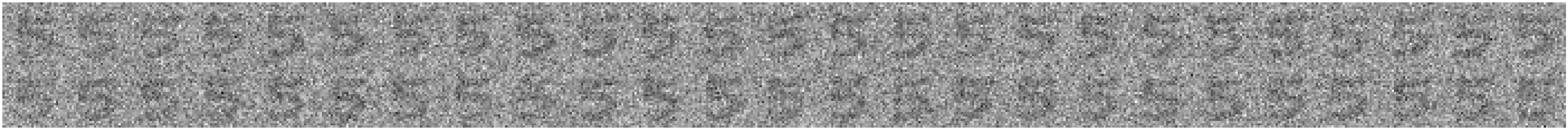}}
  \end{subfigure} 
     \begin{subfigure}[$\widehat{x}_\theta(z_{ij})$, where $z_{ij} \sim q_\phi(z|y_{ij})$ for $\sigma$-VAE trained with $\sigma=0.1$ ]{\includegraphics[width=\textwidth]{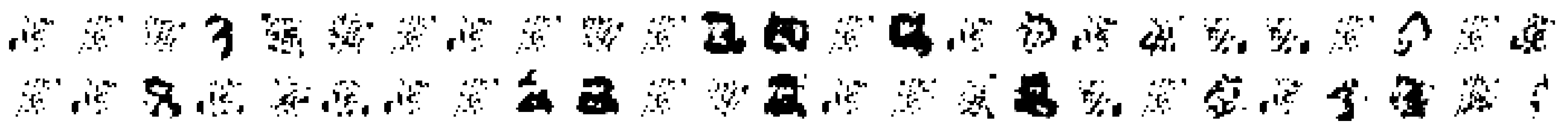}}
  \end{subfigure} 
     \begin{subfigure}[$\widehat{x}_\theta(z_{ij})$, where $z_{ij} \sim q_\phi(z|y_{ij})$ for $\sigma$-VAE trained with $\sigma=0.2$]{\includegraphics[width=\textwidth]{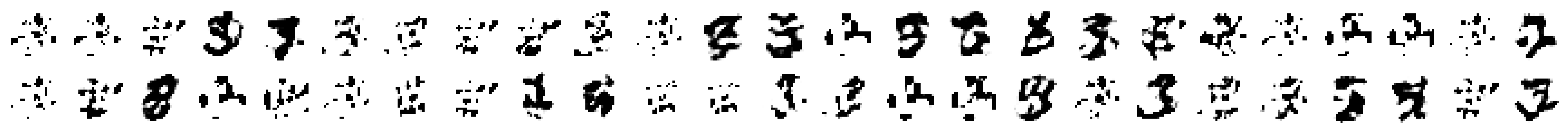}}
  \end{subfigure}   
   \begin{subfigure}[$\widehat{x}_\theta(z_{ij})$, where $z_{ij} \sim q_\phi(z|y_{ij})$ for $\sigma$-VAE trained with $\sigma=0.3$]{\includegraphics[width=\textwidth]{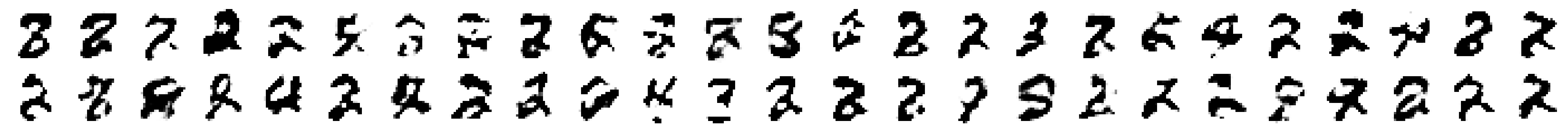}}
  \end{subfigure} 
     \begin{subfigure}[$\widehat{x}_\theta(z_{ij})$, where $z_{ij} \sim q_\phi(z|y_{ij})$ for $\sigma$-VAE trained with $\sigma=0.4$]{\includegraphics[width=\textwidth]{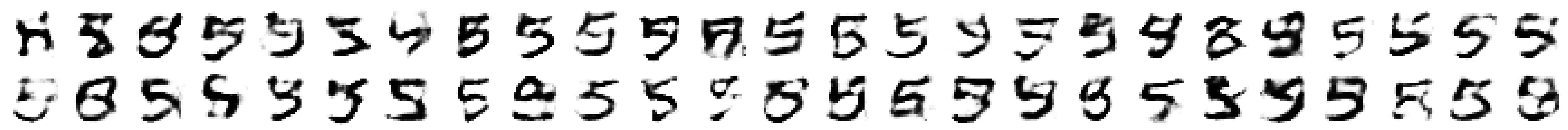}}
  \end{subfigure} 
     \begin{subfigure}[$\widehat{x}_\theta(z_{ij})$, where $z_{ij} \sim q_\phi(z|y_{ij})$ for $\sigma$-VAE trained with $\sigma=0.5$]{\includegraphics[width=\textwidth]{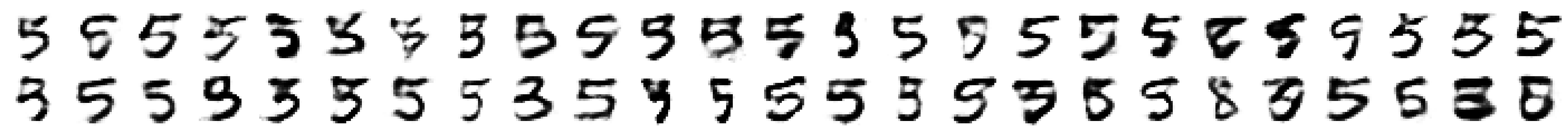}}
  \end{subfigure} 
     \begin{subfigure}[$\widehat{x}_\theta(z_{ij})$, where $z_{ij} \sim q_\phi(z|y_{ij})$ for $\sigma$-VAE trained with $\sigma=0.6$]{\includegraphics[width=\textwidth]{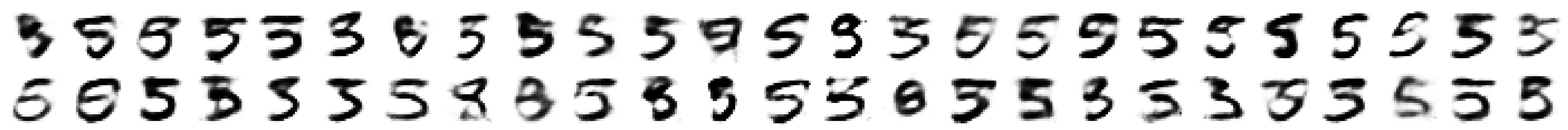}}
  \end{subfigure} 
     \begin{subfigure}[$\widehat{x}_\theta(z_{ij})$, where $z_{ij} \sim q_\phi(z|y_{ij})$ for $\sigma$-VAE trained with $\sigma=0.7$]{\includegraphics[width=\textwidth]{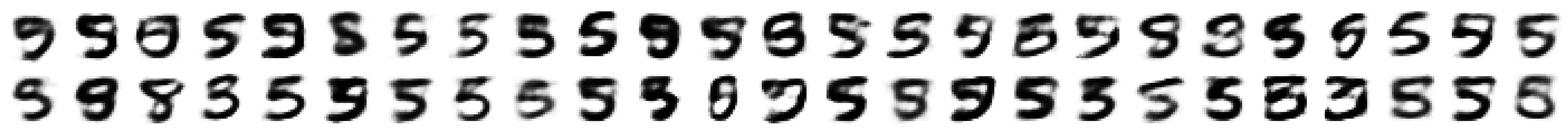}}
  \end{subfigure} 
     \begin{subfigure}[$\widehat{x}_\theta(z_{ij})$, where $z_{ij} \sim q_\phi(z|y_{ij})$ for $\sigma$-VAE trained with $\sigma=0.8$]{\includegraphics[width=\textwidth]{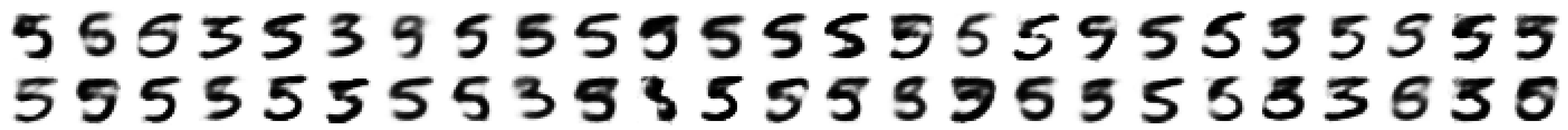}}
  \end{subfigure} 
     \begin{subfigure}[$\widehat{x}_\theta(z_{ij})$, where $z_{ij} \sim q_\phi(z|y_{ij})$ for $\sigma$-VAE trained with $\sigma=0.9$]{\includegraphics[width=\textwidth]{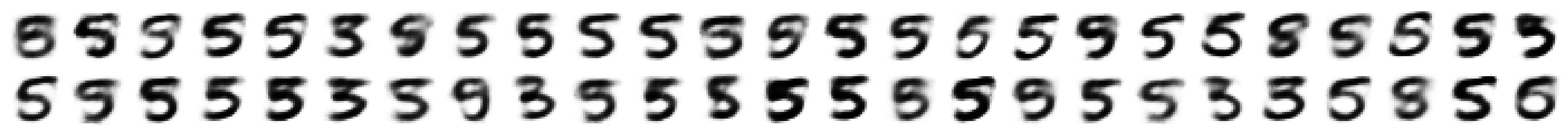}}
  \end{subfigure} 
 \caption{ The smoothness/robustness of $\sigma$-VAE is tested for models trained with different $\sigma$.}
 \end{center}
 \end{figure}

\newpage

Below we test $\sigma$-VAE vs. DEEN on high levels of salt-and-pepper noise. 

 \begin{figure}[b!]
 \begin{center} 
   \begin{subfigure}[$x_i$ from the MNIST test set]{\includegraphics[width=\textwidth]{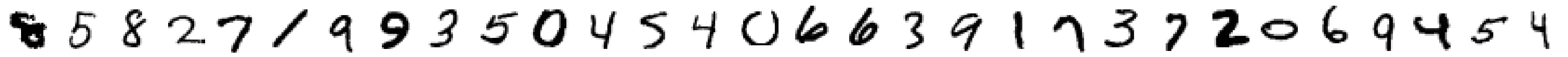}}
  \end{subfigure} 
     \begin{subfigure}[ salt-and-pepper noise with the probability $ p= 0.5$]{\includegraphics[width=\textwidth]{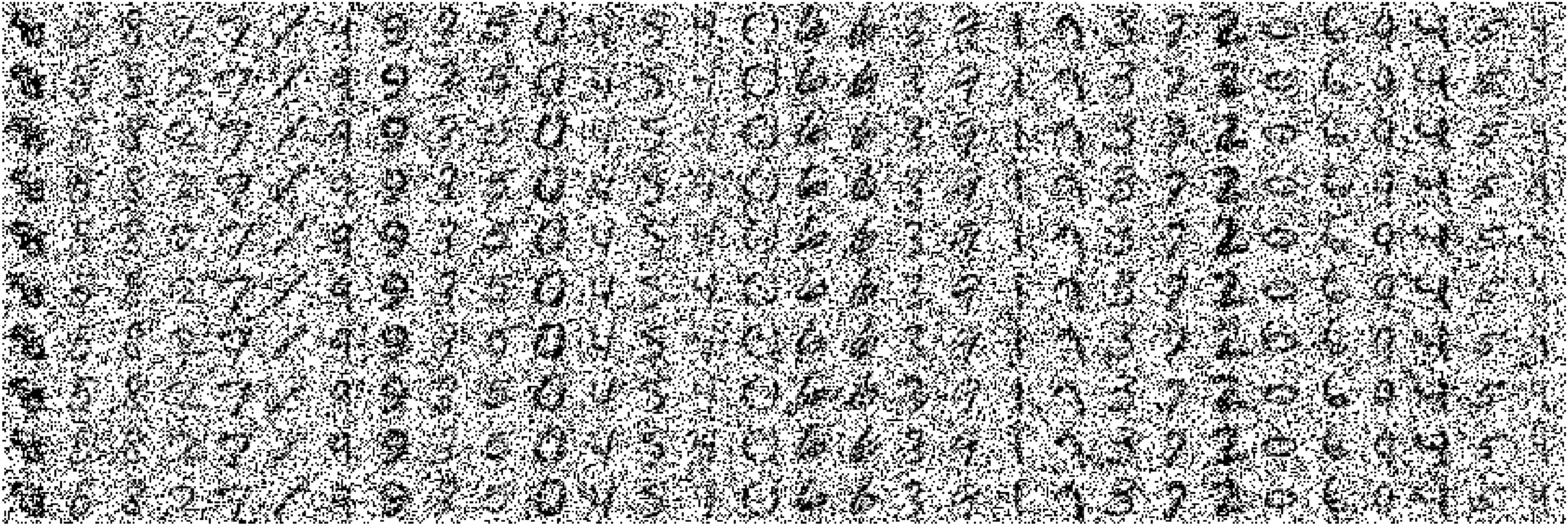}}
  \end{subfigure} 
     \begin{subfigure}[$\widehat{x}_\theta(z_{ij})$, where $z_{ij} \sim q_\phi(z|y_{ij})$ for $\sigma$-VAE trained with $\sigma=0.9$ ]{\includegraphics[width=\textwidth]{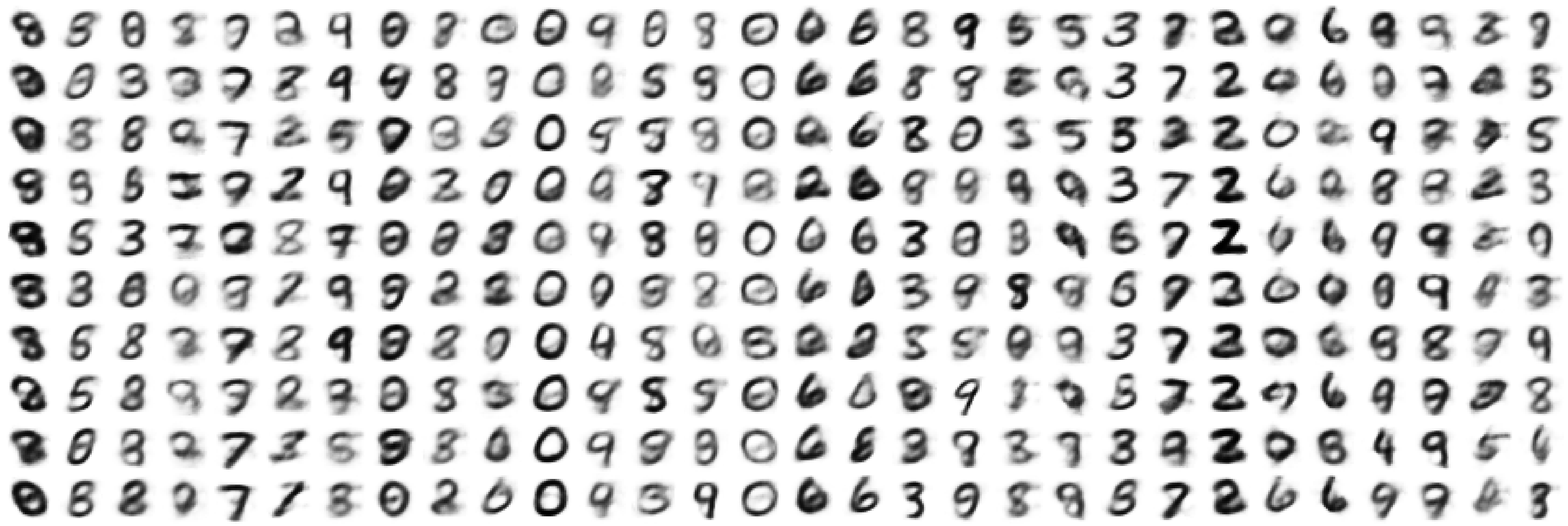}}
  \end{subfigure}   
   \begin{subfigure}[$\widehat{x}_\theta(y_{ij})=y_{ij} -\sigma^2 \nabla f_\vartheta(y_{ij})$ for DEEN trained with $\sigma=0.6$]{\includegraphics[width=\textwidth]{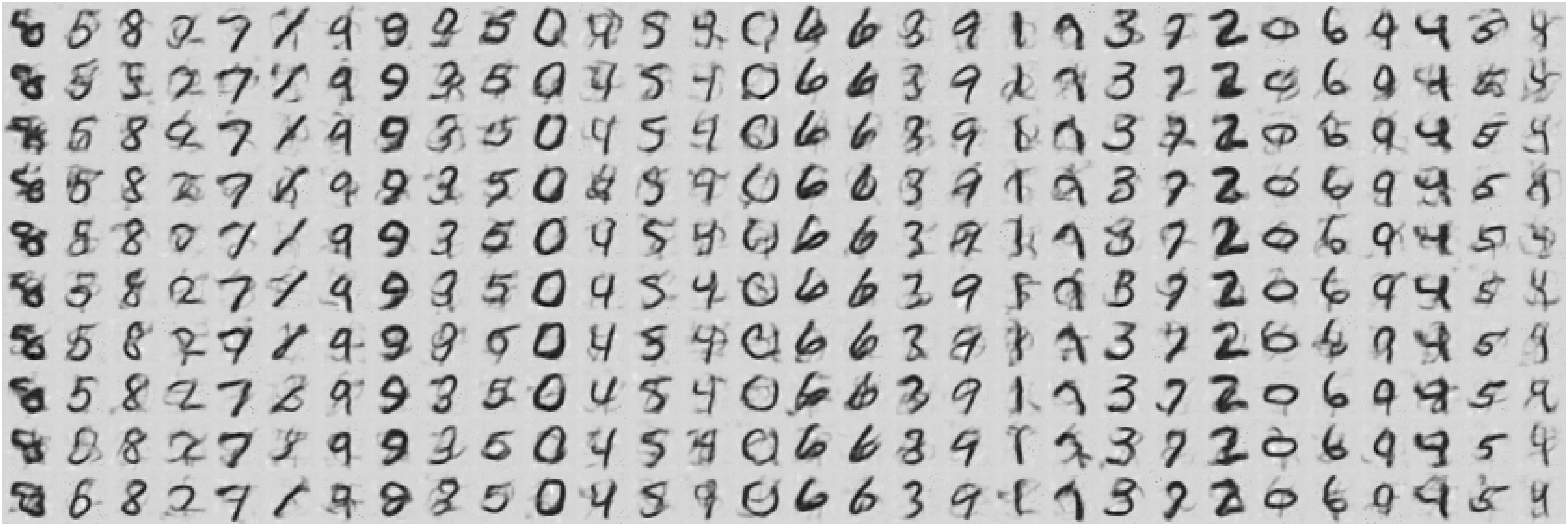}}
  \end{subfigure} 
 \caption{ (Robustness of the $\sigma$-VAE to salt-and-pepper noise.) (a) The samples $x_i$ from MNIST. (b) The samples $y_{ij}$ generated with salt-and-pepper noise with noise probability $p=0.5$. These noisy samples are held fixed for the rest. (c) The Bayes estimator $\widehat{x}_\theta(z_{ij})=\expectation[X|z_{ij}]$ is shown here, where $z_{ij} \sim q_\phi(z|y_{ij})$. The $\sigma$-VAE was the same as the previous figure: $\sigma=0.9$. Note that $\sigma$-VAE does not see any noisy samples during training. Its noise model is only imaginary. (d) The Bayes estimator $\widehat{x}_\vartheta(y_{ij})=y_{ij} -\sigma^2 \nabla f_\vartheta(y_{ij})$, where $f_\vartheta$ is learned with $\sigma=0.6$. For DEEN, denoising results were poor when we tested the model on $f_\vartheta(0.9)$, but $f_\vartheta(0.6)$ does a good job. As in the Gaussian's case, DEEN's inference-free purely density-based mechanical nature is fully visible. The vanilla VAE results trained with the same architecture are not shown here as the model simply breaks down.}
 \label{fig:bxhat}
 \end{center}
 \end{figure}

 \clearpage
 
 \section{Imaginary noise models as implicit generative models} \label{sec:appendix:samples}
 The samples $\widehat{x}(z,\theta),~ z \sim N(0,I_{d_z})$, for $\sigma$-VAE with $\sigma=0.9$ are shown below. Note that samples from the decoder's imaginary noise model are \emph{very noisy}: they are obtained by adding $\varepsilon ~\sim N(0,\sigma^2 I_d)$ to the samples shown here, therefore $z \mapsto \widehat{x}(z,\theta)$ is only an implicit generative model.
 \begin{figure}[h!]
 {\includegraphics[width=\textwidth]{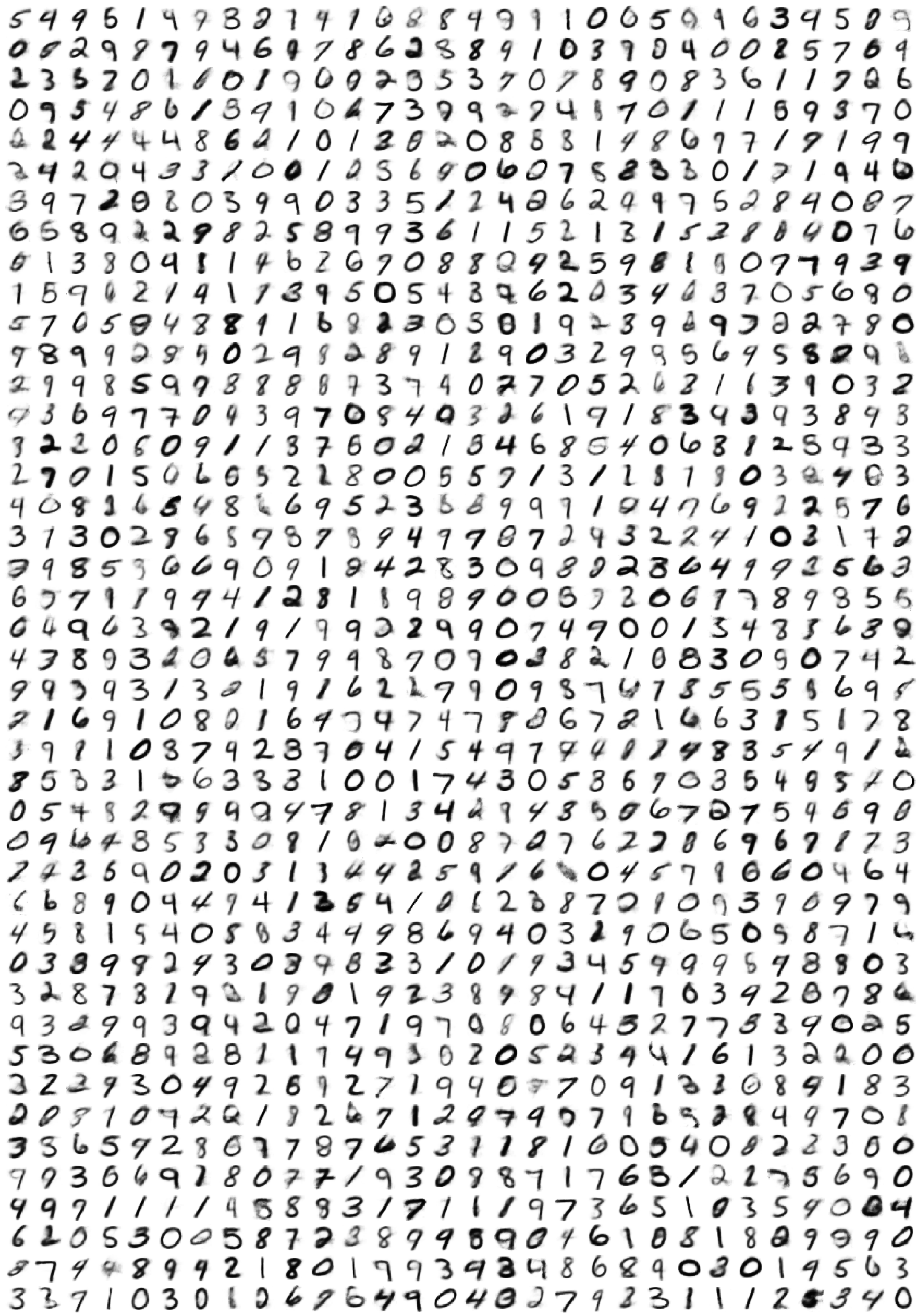}}
 \end{figure}

\end{document}